\documentclass[pdflatex,iicol,sn-mathphys-num]{sn-jnl}

\usepackage{graphicx}
\usepackage{amsmath}
\usepackage{array}
\usepackage{multirow}
\usepackage{tikz}
\usepackage{booktabs}
\usepackage{url}
\usepackage{balance}
\usepackage{amssymb}
\usepackage[linesnumbered,ruled,vlined]{algorithm2e}
\usepackage{caption}
\usepackage{comment}
\usepackage{booktabs}
\usepackage{hyperref}
\usepackage{adjustbox}
\usepackage{natbib}
\usepackage{mathpazo}

\usepackage{subcaption}
\newcolumntype{C}[1]{>{\centering\arraybackslash}p{#1}}
\newcolumntype{L}[1]{>{\raggedright\arraybackslash}p{#1}}

\theoremstyle{thmstyleone}%
\theoremstyle{thmstyletwo}%

\theoremstyle{thmstylethree}%

\usepackage{enumitem}
\setlist{itemsep=0.3em, parsep=0.2em, topsep=0.4em}

\usepackage{xcolor}
\definecolor{revisioncolor}{RGB}{0,0,0}

\newcommand{\revtext}[1]{\textcolor{revisioncolor}{#1}}

\begin{document}

\title[Article Title]{Positional Encoding in Transformer-Based Time Series Models: A Survey}


\author{\fnm{Habib} \sur{Irani}}\email{habibirani@txstate.edu}

\author*{\fnm{Vangelis} \sur{Metsis}}\email{v.metsis@txstate.edu}

\affil{\orgdiv{Computer Science Department}, \orgname{Texas State University}, \orgaddress{\city{San Marcos}, \postcode{78666}, \state{TX}, \country{USA}}}

\abstract{Recent advancements in transformer-based models have greatly improved time series analysis, providing robust solutions for tasks such as forecasting, anomaly detection, and classification. A crucial element of these models is positional encoding, which allows transformers to capture the intrinsic sequential nature of time series data. This survey systematically examines existing techniques for positional encoding in transformer-based time series models. We investigate a variety of methods, including fixed, learnable, relative, and hybrid approaches, and evaluate their effectiveness in different time series classification tasks. Our findings indicate that data characteristics like sequence length, signal complexity, and dimensionality significantly influence method effectiveness. Advanced positional encoding methods exhibit performance gains in terms of prediction accuracy, however, they come at the cost of increased computational complexity. Furthermore, we outline key challenges and suggest potential research directions to enhance positional encoding strategies. By delivering a comprehensive overview and quantitative benchmarking, this survey intends to assist researchers and practitioners in selecting and designing effective positional encoding methods for transformer-based time series models.}


\maketitle

\section{Introduction}

Time series analysis, which involves studying data collected over sequential time intervals, is fundamental across diverse domains including finance, healthcare, environmental monitoring, and industrial systems~\cite{box2015time,hyndman2018forecasting, ataei2025systematic}. The sequential nature of temporal data presents unique modeling challenges that distinguish it from other machine learning tasks, requiring specialized approaches to capture temporal dependencies, seasonal patterns, and long-range relationships effectively.


Traditional statistical methods such as Autoregressive Integrated Moving Average (ARIMA) models have long served as the foundation for time series analysis~\cite{box2015time}. While these approaches provide interpretable results and work well for stationary data with clear patterns, they face significant limitations when dealing with non-linear relationships, high-dimensional multivariate data, and complex temporal dynamics that characterize modern time series applications~\cite{hyndman2018forecasting}.
The emergence of deep learning has revolutionized time series modeling by enabling the capture of complex non-linear patterns and long-range dependencies. Recurrent Neural Networks (RNNs) and their advanced variants, including Long Short-Term Memory (LSTM) and Gated Recurrent Units (GRU), became the dominant paradigm for sequential modeling~\cite{hochreiter1997long,yamak2019comparison}. These architectures excel at processing sequential data through their inherent temporal structure, naturally handling variable-length sequences and maintaining memory of previous time steps. However, RNNs suffer from fundamental limitations including vanishing gradients, sequential processing bottlenecks, and difficulty capturing very long-range dependencies~\cite{song2018attend}.

Convolutional Neural Networks (CNNs) offer an alternative approach to time series modeling by applying convolutional filters across the temporal dimension~\cite{lecun1995convolutional,wang2017time}. While CNNs excel at capturing local temporal patterns and benefit from parallel processing, their fixed receptive fields constrain their ability to model long-term dependencies, and they require deep architectures or dilated convolutions to capture global patterns~\cite{shih2019temporal,zhao2017convolutional}.

The transformer architecture, introduced by Vaswani et al. \cite{vaswani2017attention}, has fundamentally transformed sequence modeling by replacing recurrent connections with self-attention mechanisms. This paradigm shift enables parallel processing of entire sequences while maintaining the ability to capture long-range dependencies through direct attention computation between all sequence positions \cite{li2019enhancing}. The unprecedented success of transformers in natural language processing, demonstrated by models like BERT \cite{devlin2019bert} and GPT \cite{brown2020language}, has naturally motivated their application to time series analysis \cite{zeng2023transformers,wen2023transformers,torres2021deep} with specialized architectures demonstrating superior performance across forecasting \cite{zhou2021informer,wu2021autoformer,liu2022pyraformer}, classification \cite{zerveas2021transformer,foumani2024improving}, and anomaly detection tasks \cite{xu2022anomaly,tuli2022tranad}.

The adaptation of transformers to time series has revealed both remarkable opportunities and unique challenges. Unlike discrete text tokens with semantic relationships, time series involve continuous values with complex temporal dependencies, irregular sampling patterns, and domain-specific characteristics such as seasonality, trends, and cyclical behavior \cite{li2023time, liu2023dynamic}. The inherent permutation invariance of transformer architectures, while advantageous for certain tasks, becomes problematic for time series where temporal ordering is fundamental to the data's meaning and predictive power \cite{wen2022transformers}.

This challenge has elevated positional encoding from a technical implementation detail to a critical design choice that fundamentally determines model effectiveness. \revtext{While positional encoding has been extensively studied for Natural Language Processing (NLP), time series present fundamentally different challenges. In NLP, discrete tokens possess inherent semantic relationships, and position primarily indicates word order within sentences. In contrast, time series consist of continuous values where position encodes temporal distance, phase relationships, periodicity, and causality. Time series exhibit additional complexities, including non-stationarity (statistical properties changing over time), potential irregular sampling intervals, and multiple overlapping periodicities across different timescales. These characteristics necessitate positional encoding designs specifically tailored to temporal data rather than direct adaptation of NLP methods.}

Recent comprehensive analyses have demonstrated that positional encoding methods significantly impact a model's ability to handle sequence length extrapolation, temporal pattern recognition, and cross-domain generalization \cite{zhao2024length,kazemnejad2023impact}. The unique characteristics of temporal data, including non-stationarity, multi-scale patterns, and irregular sampling, require specialized positional encoding approaches that differ substantially from those developed for natural language processing \cite{zhang2024intriguing}.

Despite extensive research in transformer architectures, the time series community lacks a comprehensive evaluation of positional encoding methods specifically designed for temporal data characteristics. Although there are general transformer surveys \cite{tay2022efficient} and NLP-focused positional encoding reviews are available \cite{zhao2024length}, no systematic study has examined how these methods perform in various time series tasks and architectures. This gap is particularly critical given the unique properties of temporal data, including nonstationarity, seasonality, and varying sampling frequencies \cite{li2023time}.

This survey addresses this critical gap by providing the first comprehensive empirical evaluation of positional encoding methods specifically for time series applications. We systematically analyze eight distinct positional encoding approaches across eleven diverse datasets representing different domains, sequence characteristics, and temporal patterns. Our evaluation spans two major transformer architectures to ensure the generalizability of findings and provides practical insights for method selection based on specific application requirements.

\revtext{This work provides the first systematic evaluation of positional encoding methods specifically for time series classification. Our key contributions include:}
\revtext{
\begin{itemize}
\item A systematic comparison of ten popular positional encoding methods across 15 diverse datasets (spanning 151 to 478,785 samples), along with their complexity analysis.
\item An evaluation of these methods across diverse time series domains on two foundational time series transformer architectures.
\item The identification of performance patterns related to sequence characteristics such as length and dimensionality.
\item Beyond empirical comparison, we analyze why certain methods excel in specific contexts, explaining TUPE's advantages on biomedical signals, SPE's effectiveness on motion data, and absolute methods' limitations on long sequences.
\item Practical guidelines for PE method selection.
\end{itemize}
}
\revtext{
While individual PE methods exist for time series and PE surveys exist for NLP, this represents the first controlled, comprehensive evaluation isolating positional encoding effects on time series classification with rigorous statistical validation across diverse domains and scales.}

Results demonstrate that advanced methods like Stochastic Positional Encoding (SPE) and Transformer with Untied Positional Encoding (TUPE) \cite{ke2020rethinking} consistently outperform traditional approaches, with effectiveness varying significantly by sequence length, dimensionality, and application domain.

The remainder of this survey is organized as follows: Section~2 provides a comprehensive literature review of transformer-based time series models and positional encoding methods. Section~3 presents background on the self-attention mechanism, positional encoding fundamentals. Section~4 details our systematic analysis of different positional encoding categories, and Sections~5-6 discuss evaluation methodologies and empirical results. 

\section{Literature Review}
This section surveys existing work on attention-based models, transformer architectures, and positional encoding methods for time series, positioning our contributions within the broader research landscape. While individual papers have proposed positional encoding variants alongside novel architectures~\cite{foumani2024improving,zhou2021informer}, and comprehensive surveys exist for NLP positional encoding~\cite{zhao2024length} and general time series transformers~\cite{wen2023transformers,torres2021deep}, no prior work has systematically isolated and evaluated positional encoding as an independent variable across diverse time series contexts. Our survey fills this gap by providing controlled, statistically rigorous comparison of positional encoding methods while holding architecture, training protocol, and evaluation methodology constant.

\subsection{Attention-based models}

The evolution of attention mechanisms in time series analysis has progressed through several distinct phases, beginning with early attention-augmented architectures and culminating in sophisticated transformer-based models specifically designed for temporal data.

Before full transformer architectures gained prominence, researchers explored attention mechanisms as enhancements to existing sequential models, building upon foundational attention work in neural machine translation \citep{bahdanau2014neural,luong2015effective}. Early work focused on augmenting traditional neural networks with attention capabilities to improve their ability to model temporal dependencies and handle variable-length sequences.

The Dual-Stage Attention-Based RNN (DA-RNN)~\cite{qin2017dual} pioneered the use of attention at both input and temporal dimensions, allowing models to selectively focus on relevant features and time steps. This approach demonstrated significant improvements in financial time series prediction by adaptively extracting relevant dependencies, establishing the foundation for attention-based temporal modeling. Similar dual attention strategies were subsequently developed for multivariate forecasting \citep{fan2019dual} and clinical time series analysis \citep{song2017attend}.

Attention-based encoder-decoder architectures adapted sequence-to-sequence learning paradigms for time series forecasting~\cite{chorowski2015attention}. These models showed that attention mechanisms could effectively handle variable-length sequences and improve long-term prediction accuracy by focusing on relevant historical information. This work demonstrated that attention could address some of the fundamental limitations of purely sequential processing approaches.

The success of self-attention motivated the development of specialized variants designed for temporal data characteristics. Cross Attention Stabilized Fully Convolutional Neural Network (CA-SFCN)~\cite{hao2020new} combined convolutional layers with two types of self-attention: temporal attention (TA) for identifying important time steps and variable attention (VA) for capturing relationships between different variables in multivariate time series. This dual attention approach showed that different attention mechanisms could be specialized for different aspects of temporal modeling.
Gated Transformer Networks (GTN)~\cite{liu2021gated} introduced two-tower multi-headed attention architectures that capture discriminative information from input series, using learnable gating mechanisms to balance between different attention patterns. This approach demonstrated that attention mechanisms could be made more adaptive to specific temporal patterns, paving the way for more sophisticated attention-based architectures.

The computational challenges of applying attention to long time series motivated the development of efficient attention mechanisms. Sparse attention patterns~\cite{beltagy2020longformer} reduce computational complexity by focusing on subsets of time steps, while maintaining the ability to capture relevant dependencies. These efficiency improvements were crucial for making attention-based approaches practical for real-world time series applications with long sequences. Complementary efficiency strategies include locality-enhanced transformers \citep{li2019enhancing} and reformer architectures \citep{kitaev2020reformer} that reduce memory bottlenecks in long sequence processing.

Recent developments have focused on adaptive attention mechanisms that adjust based on temporal context \cite{chu2021contextual,anderson2023embedding,taylor2022transformer}.

\subsection{Transformer-Based Models}

Building upon the foundation of attention mechanisms, transformer architectures represent a paradigm shift towards pure attention-based sequence modeling. In time series analysis, adaptations of Transformers have led to specialized architectures tailored to exploit their strengths while addressing temporal-specific challenges. The Transformer-based framework for multivariate time series representation learning \cite{zerveas2021transformer} integrates batch normalization to stabilize training, enhancing model performance on complex multivariate datasets. This framework underscores the significance of normalization techniques in fortifying Transformer models for diverse time series tasks. 

The multi-head attention mechanism has been extensively adapted for time series applications, with different heads specializing in various temporal patterns. The success of multi-head attention in transformers inspired specialized variants for temporal data. The Temporal Fusion Transformer (TFT) \cite{lim2021temporal} introduced specialized attention heads for different temporal patterns: some heads focus on short-term dependencies, while others capture long-term trends. The model also incorporates variable selection networks to handle high-dimensional input features common in time series applications.
Contemporary approaches have explored hierarchical and multi-scale positional encodings \cite{zhang2023hierarchical,schmidt2023multi,kumar2023positional}.

Moreover, inspired by vision Transformers, patch embeddings have been adapted for time series data. This approach, similar to the Vision Transformer (ViT) paradigm \cite{cordonnier2021differentiable}, segments time series into patches for localized processing. By encapsulating both local and global patterns, patch embeddings facilitate hierarchical pattern recognition, thereby improving model interpretability and performance in time series analysis and this type of transformer has used in several scenarios \cite{khaniki2025class, ahmadi2025unsupervised}.

However, transformers face unique challenges in time series applications. Their inherent permutation invariance necessitates explicit positional information to distinguish temporal ordering \cite{wen2023transformers, huang2020improve, zerveas2021transformer}. Additionally, the global attention mechanism may dilute focus on local patterns, and the absence of natural recency bias requires architectural adaptations for optimal time series performance.

The diversity of transformer architectures for time series reflects the heterogeneity of temporal modeling challenges across domains. Forecasting-oriented architectures like Informer~\cite{zhou2021informer}, Autoformer~\cite{wu2021autoformer}, and FEDformer~\cite{zhou2022fedformer} incorporate specialized attention mechanisms (ProbSparse, auto-correlation, frequency decomposition) that fundamentally alter how temporal information is processed. Crossformer~\cite{zhang2023crossformer} extends transformers to cross-dimensional attention for multivariate forecasting, while Pyraformer~\cite{liu2022pyraformer} employs pyramidal structures for multi-scale temporal modeling. Each architectural innovation introduces unique considerations for positional encoding: sparse attention may require efficient PE methods, auto-correlation may render certain PE approaches redundant, and frequency-domain processing may necessitate frequency-native PE representations. Our evaluation focuses on two foundational architectures (Section \ref{sec:Experimental}) to enable controlled, statistically rigorous comparison, with detailed discussion of architectural generalizability provided in Section \ref{sec:Architectural}.

\subsection{Positional Encoding in Time Series}

The adaptation of positional encoding methods to time series has revealed fundamental differences between temporal and linguistic data that require specialized approaches.

Positional encoding (PE) mechanisms are therefore crucial for enabling transformers to learn temporal relationships effectively, as self-attention mechanisms are inherently permutation-invariant. \cite{ke2021rethinking}. Vaswani et al. \cite{vaswani2017attention} initially introduced fixed sinusoidal encodings to embed positional information for temporal differentiation. While these provided a foundational approach, they can struggle with varying sequence lengths and dynamic temporal relationships common in time series data \cite{shaw2018self, lim2021temporal}. Subsequently, Shaw et al. \cite{shaw2018self} advanced the field with learnable and relative positional encodings, enhancing temporal relationship modeling capabilities.

Recent innovations have specifically targeted time series challenges. Foumani et al. \cite{foumani2024improving} developed Time Absolute Position Encoding (tAPE) and Efficient Relative Position Encoding (eRPE) for enhanced temporal dependency modeling, alongside the ConvTran model, a hybrid architecture fusing convolutional layers with transformer mechanisms. Other advanced methods include Transformer with Untied Positional Encoding (TUPE) \cite{ke2020rethinking} and Convolutional Stochastic Position Encoding (ConvSPE) \cite{liutkus2021relative}, which address content-position decoupling and linear complexity maintenance, respectively. Hybrid approaches, combining absolute and relative strategies, have also been proposed to leverage complementary strengths for comprehensive temporal information representation, aiming to address some of the inherent limitations of transformer architectures in time series analysis.

Recent work has systematically evaluated PE methods for time series applications \cite{alioghli2025enhancing,martinez2023positional}. Advanced techniques include frequency-based encodings \cite{wang2023frequency,nguyen2023spectral}, learnable approaches \cite{li2023learnable,foster2023learning}, and hybrid strategies \cite{miller2023hybrid,bell2023adaptive}. Specialized methods for irregular time series \cite{campbell2023transformer,lee2022adaptive} and long sequences \cite{patel2023efficient,chen2023adaptive} have also been developed.

\begin{table*}[t]
    \centering
    \caption{\revtext{\small Summary of Positional Encoding Techniques}}
    \label{tab:techniques}
    \fontsize{8}{9.6}\selectfont
    \setlength{\tabcolsep}{6pt}
    \renewcommand{\arraystretch}{1.15}
    \begin{tabular}{@{}L{2cm}p{1.4cm}L{4.0cm}p{2.5cm}p{0.9cm}@{}}
    \toprule
    \textbf{Method} & \textbf{Type} & \textbf{Injection Technique} & \textbf{Learnable/Fixed} & \textbf{Ref.} \\ \midrule
    Sin. PE & Absolute & Additive Embedding & Fixed & \cite{vaswani2017attention} \\
    Learn. PE & Absolute & Additive Embedding & Learnable & \cite{vaswani2017attention} \\
    RPE & Relative & Attention Manipulation & Fixed & \cite{shaw2018self} \\
    tAPE & Absolute & Additive Embedding & Fixed & \cite{foumani2024improving} \\
    RoPE & Hybrid & Attention Manipulation & Fixed & \cite{su2024roformer} \\
    eRPE & Relative & Attention Manipulation & Learnable & \cite{foumani2024improving} \\
    TUPE & Hybrid & Attention Manipulation & Learnable & \cite{ke2020rethinking} \\
    ConvSPE & Relative & Attention Manipulation & Learnable & \cite{liutkus2021relative} \\
    T-PE & Hybrid & Combined Technique & Mixed & \cite{zhang2024intriguing} \\
    ALiBi & Relative & Attention Manipulation & Fixed & \cite{press2021train} \\
    \bottomrule
    \end{tabular}
\end{table*}

Table \ref{tab:techniques} provides a comprehensive taxonomy of the positional encoding techniques discussed, based on criteria such as technique type, injection method, learning strategy, parameters, and time complexity. The parameter counts include all learnable parameters across layers, where $l$ represents the number of transformer layers, $h$ is the number of attention heads, and $d$ is the hidden dimension. For methods using relative positioning (RPE, eRPE), we include the full attention matrix parameters. For ConvSPE, K represents the convolutional kernel size that controls the local neighborhood for convolution operations.

\revtext{
\textbf{Emerging and Specialized Positional Encoding Approaches:} 
Recent research has explored positional encoding strategies beyond the sequence-based methods evaluated in this survey, targeting specialized scenarios and data structures.}

\revtext{
\textbf{Graph-Based Positional Encoding:} For multivariate time series with known variable relationships, graph-based approaches encode both temporal position and cross-variable dependencies. GRformer~\cite{grformer2024} uses Graph Neural Networks with RNN-based positional embedding generation to model inter-variable correlations in forecasting tasks. Similarly, dynamic graph convolutional methods~\cite{chen2024ddgct} construct adaptive spatial-temporal graphs for anomaly detection. While promising, these approaches require explicit specification of variable relationship graphs, which limits applicability to datasets without prior domain knowledge of inter-variable dependencies.}

\revtext{
\textbf{Task-Specific Positional Encoding:} Certain PE methods are designed for specific tasks or architectures. Continuous positional encoding~\cite{guo2022stctn} links encoder and decoder positions in forecasting models, maintaining positional continuity across the historical-future boundary. Context-aware positional encoding~\cite{cheng2024ctnet} adapts representations based on local sequence characteristics, enhancing temporal invariance for classification tasks with irregular patterns. These specialized methods demonstrate strong performance in their target domains but require architectural modifications that limit their applicability across diverse transformer variants.}

\revtext{
\textbf{Scope of This Survey:} Our evaluation focuses on sequence-based positional encoding methods that: (1) operate on temporal sequences without requiring additional graph structures, (2) are applicable to classification tasks, (3) can be integrated into standard transformer architectures without major modifications, and (4) have sufficient maturity and documentation for rigorous benchmarking. Graph-based methods, while important for specific multivariate scenarios, represent a complementary research direction requiring different experimental protocols and data prerequisites. Similarly, forecasting-specific methods (e.g., continuous PE) are optimized for encoder-decoder architectures that differ from the classification-focused transformers in our evaluation. Future surveys should examine these specialized approaches within their respective contexts.}

\section{Positional Encoding Analysis}
This section offers a fundamental explanation of self-attention and a summary of existing positional encoding models. It's important to distinguish between positional encoding, which refers to the approach used to incorporate positional information (e.g., absolute or relative), and positional embedding, which denotes the numerical vector associated with that encoding.


\subsection{Problem definition and notation}
Given a time series dataset \( X \) consisting of \( n \) samples, \( X = \{x_1, x_2, \dots, x_n\} \), where each sample \( x_i = \{x_{i1}, x_{i2}, \dots, x_{iL}\} \) represents a \( d_x \)-dimensional time series of length \( L \), such that \( x_i \in \mathbb{R}^{L \times d_x} \). Additionally, let \( Y = \{y_1, y_2, \dots, y_n\} \) denote the corresponding set of labels, where \( y_t \in \{1, 2, \dots, c\} \) and \( c \) is the number of classes. The objective is to train a neural network classifier to map the input dataset \( X \) to its corresponding label set \( Y \).

\subsection{Self-Attention Mechanism}


\begin{figure}[t]
    \centering
    \includegraphics[
        width=\columnwidth,
        trim=1.5cm 0cm 0cm 0cm,
        clip
    ]{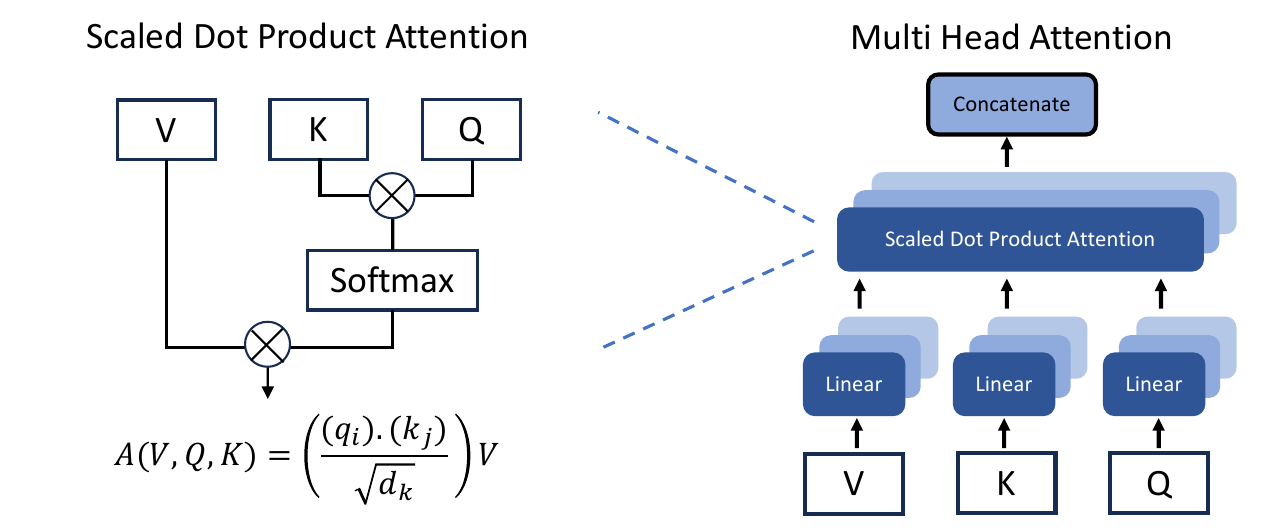}
    \caption{\small Scaled dot-product attention (left) and multi-head attention (right) mechanisms in transformer architecture.}
    \label{fig:self_attention}
\end{figure}

The self-attention mechanism is a key component of Transformer models, enabling them to capture long-range dependencies in sequential data. Given an input sequence \( X \in \mathbb{R}^{L \times d_x} \), the self-attention mechanism computes a weighted representation of each element in the sequence by attending to all other elements, as illustrated in Fig. \ref{fig:self_attention}.

First, three linear projections are applied to the input sequence to obtain the query (\(Q\)), key (\(K\)), and value (\(V\)) matrices:

{\small
\[
Q = XW_Q, \quad K = XW_K, \quad V = XW_V,
\]
}


followed by scaled dot-product attention:
{\small
\[
Attention(Q, K, V) = softmax\left(\frac{QK^\top}{\sqrt{d_k}}\right)V,
\]
}

The output of the self-attention mechanism is a weighted sum of the value vectors, where the weights are determined by the attention scores. This process allows the model to focus on the most relevant parts of the sequence when generating each element of the output.

The theoretical foundations of attention mechanisms have been extensively analyzed \cite{yun2020are,tsai2019transformer}, revealing their universal approximation capabilities for sequence-to-sequence functions. Recent work has provided deeper insights into attention patterns and their interpretability \cite{clark2019what}, which is particularly relevant for understanding how different positional encodings influence attention weight distributions in time series contexts.


Multi-head attention concatenates parallel attention heads to learn diverse representations:
{\footnotesize
\begin{align*}
\text{MultiHead}(Q, K, V) &= \text{Concat}(\text{head}_1, \text{head}_2, \dots, \text{head}_h) W_O
\end{align*}
}

Transformers lack an inherent notion of sequence order, as self-attention operates on sets of inputs without considering their positions. To address this, positional encoding is introduced to incorporate positional information into the model, allowing it to understand the order of elements in a sequence.

Let the input sequence be \( X \in \mathbb{R}^{L \times d_x} \), where \( L \) is the sequence length and \( d_x \) is the feature dimension. Positional encoding adds a unique vector to each position in the sequence, resulting in a new representation \( X' \):
\vspace{-3mm}
\[
X' = X + P,
\]
where \( P \in \mathbb{R}^{L \times d_x} \) is the positional encoding matrix.

The theoretical foundations of positional encoding in sequential models have been extensively analyzed from both signal processing \cite{sun2023convolutional,kim2023temporal} and cybernetic control perspectives \cite{peters2022deep,thompson2023cybernetic}.

\subsection{Absolute Positional Encoding}
Absolute positional encoding assigns unique vectors to each temporal position. The sinusoidal approach \cite{vaswani2017attention} provides computational efficiency by using predetermined trigonometric functions, eliminating the need for learned positional representations.
\subsubsection{Fixed Absolute Positional Encoding}
The fixed absolute positional encoding is defined as follows:
\vspace{-3mm}
{\small
\begin{align*}
    PE_{(pos, 2i)} &= \sin\left(\frac{pos}{10000^{\frac{2i}{d}}}\right) \\
    PE_{(pos, 2i+1)} &= \cos\left(\frac{pos}{10000^{\frac{2i}{d}}}\right)
\end{align*}
}
where \( pos \) is the position (time step) in the sequence, \( i \) is the dimension, and \( d \) is the dimensionality of the positional encoding.

The use of sinusoidal functions ensures that the encodings for nearby positions are similar, which can help the model learn relationships between closely spaced time steps. However, this approach has limitations when dealing with sequences of varying lengths or non-uniform time intervals, which are common in time series data.

\subsubsection{Learnable Positional Encoding}
Learnable positional encoding provides more flexibility by allowing the model to learn the positional representations during training. This can be advantageous for time series tasks where the relationships between time steps may be complex or nonlinear. The learnable positional encoding is typically defined as:

\begin{equation*}
PE_{learnable}({pos}) = W_{{pos}}
\end{equation*}
where \( W_{pos} \) is an \( d_{\text{model}} \)-dimensional vector representing the learnable positional encoding parameters for position \( pos \).

During the forward pass of the neural network, the learnable positional encoding is added to the input embeddings \( x \):
\begin{equation*}
x = x + PE_{learnable}[:x.size(1), :]
\end{equation*}
where \( x \) represents the input tensor at a given position and \( x.\text{size}(1) \) denotes the length of the sequence up to the current position. This addition enables the model to learn position-specific transformations that can capture dependencies and patterns within sequences.

The flexibility and adaptability of learnable positional encoding make it a powerful tool for enhancing the modeling of temporal dependencies and patterns in time series data within transformer architectures.

Then, the self-attention mechanism is modified as follows:
{\footnotesize
\begin{equation*}
    \text{Attention}(Q, K, V) = \text{softmax}\left(\frac{(Q + PE) \cdot (K + PE)^T}{\sqrt{d_k}}\right) V
\end{equation*}
}
where \( Q \), \( K \), and \( V \) are the query, key, and value matrices, respectively, \( PE \) represents the positional encoding matrix, and \( d_k \) is the dimensionality of the key vectors \cite{vaswani2017attention}.

\subsection{Relative Positional Encoding}

Shaw et al. \cite{shaw2018self} proposed modifying the attention computation to include relative position information:


\vspace{-3mm}
{\small
\begin{equation*}
    e_{ij} = \frac{(x_i W_Q) (x_j W_K + a^K_{ij})^T}{\sqrt{d_z}}
\end{equation*}

\begin{equation*}
    e_{ij} = \frac{(x_i W_Q) (x_j W_K)^T + (x_i W_Q) (a^K_{ij})^T}{\sqrt{d_z}}
\end{equation*}

\begin{equation*}
    a^K_{ij} = W_K^r r_{i-j}
\end{equation*}
}
where \( x_i \) and \( x_j \) are the input representations at positions \( i \) and \( j \),  
\( W_Q \) and \( W_K \) are the learnable projection matrices for queries and keys,  
and \( a^K_{ij} \) is the relative positional encoding defined as \( a^K_{ij} = W_K^r r_{i-j} \),  
where \( r_{i-j} \) is the learnable relative positional embedding for the distance \( i - j \).

This formulation allows the model to better capture temporal dependencies that depend on the relative position between time steps, which is particularly useful for tasks such as time series forecasting or classification when dealing with varying sequence lengths \cite{shaw2018self, raffel2020exploring}.

\begin{figure*}[t]
    \centering
    \includegraphics[width=0.85\textwidth,trim={0cm 0 0 0},clip]{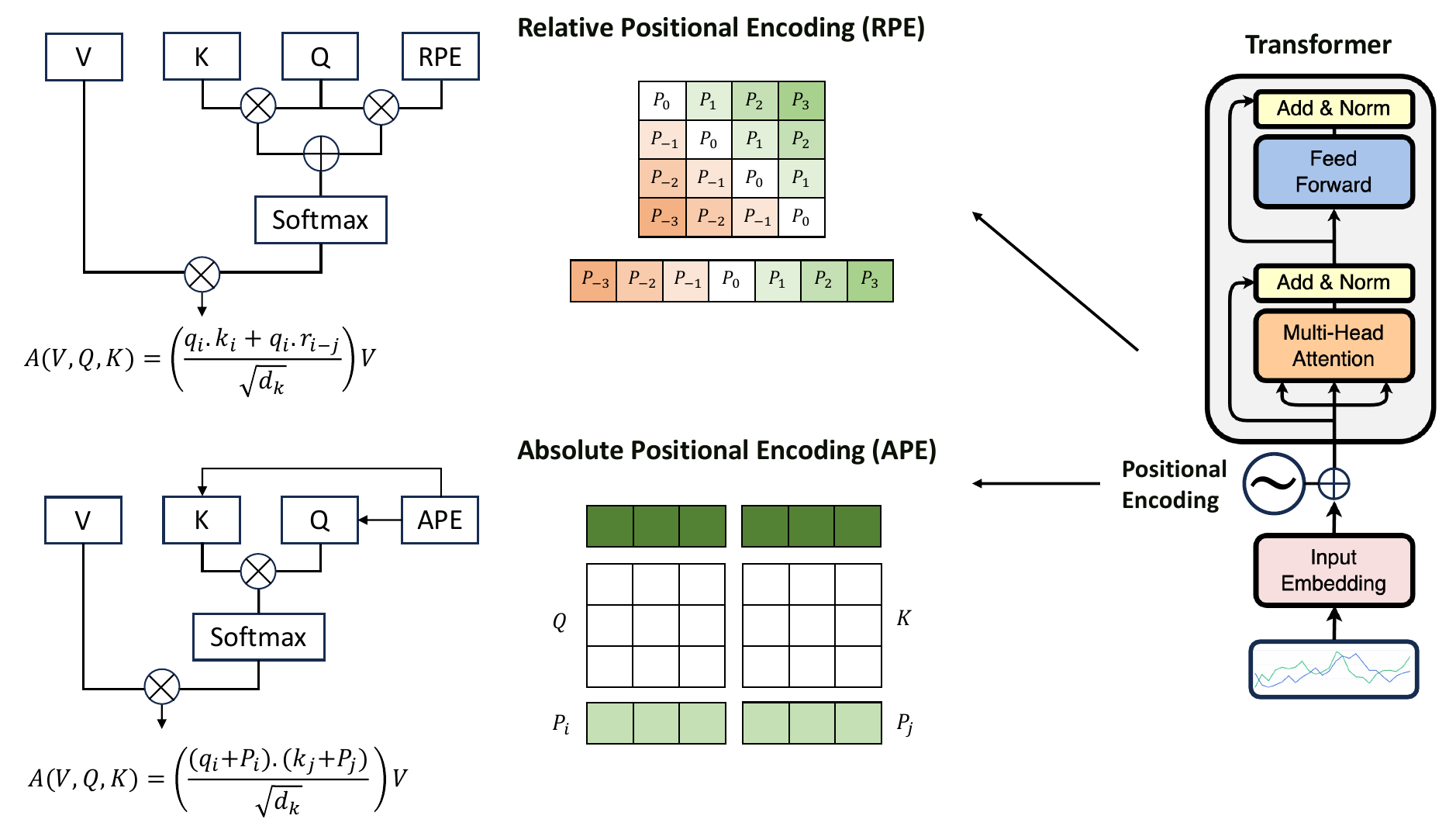}
    \caption{\revtext{\small Absolute vs. relative positional encoding: absolute methods add position vectors to input embeddings, while relative methods directly incorporate positional relationships in attention computation.}}
    \label{fig:positional_encodings}
    \vspace{-3mm}
\end{figure*}

\begin{figure}
    \centering
    \includegraphics[width=0.98\columnwidth,clip]{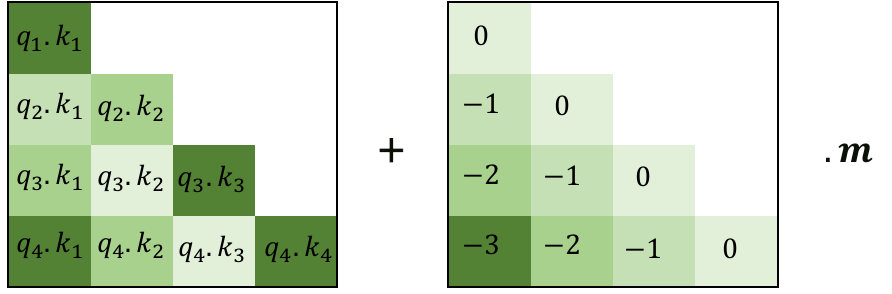}
    \caption{\small The left matrix represents standard attention score computation \(q_i \cdot k_j\) between queries and keys. The right matrix shows ALiBi's linear bias values that are added to attention scores before softmax normalization.}
    \label{fig:Hybrid}
\end{figure}

As illustrated in Fig. \ref{fig:Hybrid}, positional encoding methods can be combined to leverage the advantages of both absolute and relative approaches.

\subsection{Time Absolute Position Encoding (tAPE)}

Foumani et al. \cite{foumani2024improving} identified that traditional sinusoidal encoding loses distance awareness in lower-dimensional embeddings. tAPE addresses this by incorporating sequence length into frequency computation:

\vspace{-5mm}
\begin{equation*}
\omega_k^{new} = k \times \frac{d_{model}}{L}
\end{equation*}
\begin{equation*}
PE_{pos,2i} = \sin(pos \times \omega_k{new})
\end{equation*}
\begin{equation*}
PE_{pos,2i+1} = \cos(pos \times \omega_k{new})
\end{equation*}

This modification ensures consistent distance awareness across different embedding dimensions and sequence lengths, making it particularly suitable for diverse time series applications.

\subsection{Rotary Positional Encoding (RoPE)}

Rotary Positional Encoding (RoPE)~\cite{su2024roformer} uses rotation matrices to encode absolute positions while incorporating explicit relative position dependencies. This method applies rotations to query and key representations based on their absolute positions before computing dot-product attention:

\begin{equation}
e_{ij} = \frac{(x_i W^Q + R_{n\theta})(x_j W^K + R_{n\theta})^T}{\sqrt{d_z}}
\end{equation}

where $R_{n\theta}$ is a rotation matrix that rotates by $n\theta$ radians:

\begin{equation}
R_{n\theta} = \begin{bmatrix}
\cos(n\theta) & -\sin(n\theta) \\
\sin(n\theta) & \cos(n\theta)
\end{bmatrix}
\end{equation}

where $n$ is the position of the data, and $\theta$ is a scalar value (angle) chosen to rotate the value. RoPE combines the benefits of both absolute and relative PE, ensuring that attention calculations depend on relative distances while maintaining absolute position information. This hybrid approach has shown particular effectiveness in capturing long-range dependencies in sequential data.

\subsection{Efficient Relative Position Encoding (eRPE)}





This method enhances traditional relative positioning by applying bias terms after the softmax operation:

\vspace{-3mm}
{\small
\begin{equation*}
\alpha_i = \sum_{j\in L} \left(\underbrace{\frac{\exp(e_{i,j})}{\sum_{k\in L} \exp(e_{i,k})}}{A{i,j}} + w_{i-j}\right)x_j
\end{equation*}
where:
}


where $w \in \mathbb{R}^{2L-1}$ represents learnable relative position biases. The post-softmax application sharpens positional distinctions, making the method particularly effective for tasks requiring precise temporal localization.

\revtext{\subsection{Transformer with Untied Positional Encoding (TUPE)}}

\revtext{
Ke et al.~\cite{ke2020rethinking} observe that traditional positional encoding methods add positional information directly to content embeddings ($X = X_{\text{content}} + X_{\text{pos}}$), which forces the model to process both types of information through the same transformation matrices. This can be problematic when content and position require different representational transformations.}

\revtext{
TUPE addresses this by maintaining separate pathways for content and positional information. Rather than combining them before attention computation, TUPE applies distinct transformations:}

\revtext{
\begin{align}
Q_c = X_{\text{content}} W_Q^c, \quad K_c &= X_{\text{content}} W_K^c \\
Q_p = X_{\text{pos}} W_Q^p, \quad K_p &= X_{\text{pos}} W_K^p
\end{align}
}

\revtext{
The attention score is then computed as a sum of four interaction terms:}

\revtext{
\begin{equation}
\begin{split}
A_{ij} = \frac{1}{\sqrt{d}}(&Q_c^{(i)} \cdot (K_c^{(j)} + K_p^{(j)}) \\
&+ Q_p^{(i)} \cdot (K_c^{(j)} + K_p^{(j)}))
\end{split}
\end{equation}
}

\revtext{
This decomposition allows the model to learn different types of relationships: $Q_c \cdot K_c$ captures content-based similarity, $Q_p \cdot K_p$ encodes pure positional relationships, while the cross terms $Q_c \cdot K_p$ and $Q_p \cdot K_c$ model interactions between content and position.}
\revtext{
The separation addresses a fundamental challenge in joint content-position encoding. When using a single transformation matrix, the model must balance between extracting meaningful content features and preserving positional information—objectives that may require conflicting weight configurations. For instance, in physiological signal analysis, identifying waveform morphology (content) requires different feature extractors than encoding inter-event timing (position). By separating these pathways, TUPE allows each to be optimized more independently. During backpropagation, the gradient with respect to content parameters becomes $\frac{\partial \mathcal{L}}{\partial W_Q^c} = \frac{\partial \mathcal{L}}{\partial Q_c}X_c^T$, which primarily responds to content-related patterns, while $\frac{\partial \mathcal{L}}{\partial W_Q^p} = \frac{\partial \mathcal{L}}{\partial Q_p}X_p^T$ primarily responds to positional patterns, reducing the interference that occurs when both signals pass through shared weights.}

\revtext{
This architectural choice is particularly beneficial for time series, where temporal patterns often exhibit distinct content and positional characteristics. Content features may capture local patterns like amplitude variations or shape characteristics, while positional patterns encode longer-range dependencies such as periodic cycles or seasonal trends. TUPE's separate pathways enable specialized modeling of these heterogeneous patterns.}

\revtext{
The increased parameter count (four weight matrices instead of two) results in approximately 30\% more parameters than standard transformers, but empirical results suggest this cost is justified by improved representation learning, particularly for complex sequences with rich positional structure.}





\revtext{\subsection{Stochastic Positional Encoding (SPE)}}






\revtext{
Liutkus et al.~\cite{liutkus2021relative} propose Stochastic Positional Encoding (SPE), which replaces the computationally expensive relative position matrices with convolutional operations. The core idea is to leverage convolution's inherent properties to encode relative distances efficiently.
Given input sequence $X \in \mathbb{R}^{L \times d}$, SPE computes attention scores by adding a convolutional positional bias:}

\revtext{
\begin{equation}
A_{i,j} = \frac{Q_i \cdot K_j^T}{\sqrt{d}} + \text{Conv1D}(r, i-j)
\end{equation}
}

\revtext{where $r$ is a learnable positional embedding and $\text{Conv1D}$ is a 1D convolution with kernel size $K$.}

\revtext{
The effectiveness of convolution for positional encoding stems from its translation equivariance property. When a convolution is applied to a sequence shifted by $\tau$ positions, the output is also shifted by $\tau$ positions. This means the positional bias between any two tokens depends only on their relative distance $i-j$, not their absolute positions, precisely the behavior required for relative positional encoding. Additionally, the limited kernel size $K$ naturally restricts attention to nearby positions, providing an inductive bias that aligns well with temporal locality in time series, where recent timesteps typically have stronger influence than distant ones.
}

\revtext{
To further reduce computational cost, SPE employs stochastic approximation through random Fourier features. Instead of maintaining explicit embeddings for all possible relative distances, SPE approximates the positional function as:}

\revtext{
\begin{equation}
r_\tau \approx \sum_{k=1}^{R} \alpha_k \phi_k(\tau)
\end{equation}
}

\revtext{
where $\phi_k(\tau)$ are fixed random basis functions and $\alpha_k$ are learned coefficients, with $R \ll L$. This approximation exploits the observation that meaningful positional information often has low intrinsic dimensionality. Positional relevance typically decays smoothly with distance and exhibits periodic patterns that can be captured efficiently with a small number of Fourier components. This reduces memory complexity from $\mathcal{O}(Ld)$ to $\mathcal{O}(Rd)$ while maintaining accuracy, as the random feature approximation converges rapidly with increasing $R$.}

\revtext{
The combination of convolutional structure and stochastic approximation gives SPE a computational complexity of $\mathcal{O}(LKR)$ compared to $\mathcal{O}(L^2d)$ for traditional relative positional encodings, making it particularly suitable for long time series sequences.
}

\subsection{Temporal Positional Encoding (T-PE)}




This method combines geometric and semantic components to capture both positional and content-based temporal relationships:
The semantic component adjusts attention scores by incorporating token similarity between positions \(i\) and \(j\):
\vspace{-3mm}
{\small
\begin{equation*}
T\text{-}PE(i) = PE(i) + S(i, j)
\end{equation*}

\begin{equation*}
S(i, j) = \exp\left(-\frac{\|x_i - x_j\|^2}{2\sigma^2}\right)
\end{equation*}
}
The geometric component maintains traditional positional information, while the semantic component $S(i,j)$ measures similarity between tokens at different positions, enabling the model to identify recurring patterns regardless of their absolute positions.

\subsection{Attention with Linear Biases (ALiBi)}

This method ~\cite{press2021train} introduces a fundamentally different approach by eliminating traditional positional embeddings. Instead, ALiBi directly modifies attention scores by adding linear biases proportional to the distance between query and key positions:

\begin{equation*}
\text{Attention}(Q, K, V) = \text{softmax}\left(\frac{QK^T}{\sqrt{d_k}} + \mathbf{m} \cdot \mathbf{B}\right)V
\end{equation*}

where $\mathbf{B}$ is the bias matrix with elements $B_{ij} = -|i - j|$ representing the absolute distance between positions $i$ and $j$, and $\mathbf{m}$ is a head-specific slope parameter controlling bias strength. The slope values are computed using a geometric sequence:

\begin{equation*}
m_h = 2^{-\frac{8h}{H}} \text{ for } h = 1, 2, \ldots, H
\end{equation*}

For time series applications, ALiBi uses symmetric distance calculations $B_{ij} = -|i - j|$ to enable bidirectional attention across temporal sequences. This parameter-free approach naturally encodes local temporal pattern preference while permitting long-range dependencies and enables extrapolation to longer sequences than seen during training.

\section{Experimental Setup}
\label{sec:Experimental}

The experimental setup involves comparing different positional encoding methods in two distinct Transformer-based architectures for time series data. The Time Series Transformer \cite{zerveas2021transformer}, processes the raw multivariate time series data directly. For a time series with \(L\) timesteps and \(d\) channels, it treats each timestep as a token, resulting in \(L\) tokens of dimension \(d\) that are fed into the transformer layers. This architecture incorporates normalization layers after both the attention and feed-forward components to stabilize training and address internal covariate shifts. In contrast, the patchTST \cite{cordonnier2021differentiable}, first transforms the input data using convolutional operations to create overlapping patches. Each patch spans multiple timesteps but maintains the full channel dimension, effectively reducing the sequence length while enriching the feature representation of each token. This results in \(L/p\) tokens (where \(p\) is the patch size) with increased feature dimensionality before being processed by the transformer layers. These architectures were chosen to evaluate positional encoding methods under different sequence processing paradigms: direct temporal processing in the Time Series Transformer versus hierarchical feature extraction through patch embedding. The complete code implementation and benchmarks are made publicly available for reproducibility: {\small \url{https://github.com/imics-lab/positional-encoding-benchmark}}.

\subsection{System Configuration}

We conducted all experiments on a high-performance Linux server with the following specifications: an AMD EPYC 7513 32-Core Processor (2 sockets, 128 threads), one NVIDIA RTX A5000 GPUs (24GB GDDR6 each, CUDA version 12.2, Driver version 535.113.01), 503GB of RAM, and a dual storage system consisting of an 892.7GB NVMe SSD and a 10.5TB HDD. The operating system used was Ubuntu 20.04.6 LTS (Kernel 5.4.0-200-generic). All implementations were carried out using Python 3.8.10 and PyTorch 2.4.1+cu121.

\subsection{Datasets} A comprehensive set of time series datasets from various domains is employed to evaluate the positional encoding methods' effectiveness across diverse applications. The selected datasets include:

\begin{itemize} 
\item \textbf{Sleep:} EEG (Electroencephalogram) signals with 178 time steps per instance. The dataset contains 478,785 training instances and 90,315 test instances, classified into five different stages of sleep. 
\revtext{\item \textbf{CardiacArrhythmia (CA):} Single-lead ECG recordings from the PhysioNet/Computing in Cardiology Challenge 2017 \cite{clifford2017af,goldberger2000physiobank}, with 1,500 time steps per instance (representing 5 seconds at 300Hz sampling rate). The dataset contains 43,641 training instances and 10,661 test instances across three classes: normal sinus rhythm, atrial fibrillation, and other cardiac rhythms. Data was collected using AliveCor mobile ECG devices.}
\revtext{\item \textbf{InsectSound (IS):} Audio time series capturing wingbeat patterns of flying insects \cite{chen2014flying,dau2019ucr}. Each instance contains 600 time steps (10ms segments sampled at 6,000Hz) representing amplitude-modulated signals from infrared beam occlusion. The dataset has 25,000 training and 25,000 test instances across 10 balanced classes, including male and female specimens of various mosquito species (Aedes, Culex Quinquefasciatus, Culex Stigmatosoma, Culex Tarsalis), fruit flies, and house flies.}
\item \textbf{ElectricDevices:} Electrical consumption data of household devices with 96 time steps. It contains 8,926 training instances and 7,711 test instances across seven device classes \cite{UCRArchive2018}. 
\item \textbf{FaceDetection:} EEG signals collected during face detection tasks, with 62 time steps per instance. The dataset has 5,890 training instances and 3,524 test instances for a binary classification task. 
\item \textbf{MelbournePedestrian:} Traffic data capturing pedestrian flow in Melbourne. Each instance contains 24 time steps, with a 10-class classification task, comprising 1,194 training instances and 2,439 test instances. 
\item \textbf{SharePriceIncrease:} Financial time-series data with 60 time steps per instance, used to predict whether a company's share price will increase. The dataset has 965 training instances and 965 test instances, with two classes. 
\item \textbf{LSST:} Astronomical time-series data from the Large Synoptic Survey Telescope, containing 36 time steps per instance. The dataset includes 2,459 training instances and 2,466 test instances for a 14-class classification task of celestial objects. 
\item \textbf{RacketSports:} Human activity recognition data from racket sports like tennis and badminton. Each instance contains 30 time steps, with 151 training instances and 152 test instances across four activity classes. 
\item \textbf{SelfRegulationSCP1:} EEG data related to self-regulation through slow cortical potentials (SCPs), consisting of 896 time steps per instance. The dataset contains 268 training instances and 293 test instances for binary classification. 
\revtext{\item \textbf{SelfRegulationSCP2 (SR2):} Another EEG-based self-regulation dataset with 6 channels and longer sequences \cite{birbaumer1999slow,blankertz2004bci}, containing 1,152 time steps per instance. The dataset has 200 training instances and 180 test instances for binary classification. This dataset presents additional challenges due to its extended sequence length and subtle signal patterns.}
\revtext{\item \textbf{JapaneseVowels (JV):} Audio time series with 12 LPC (Linear Predictive Coding) cepstrum coefficients recorded from nine male speakers \cite{kudo1999japanese,uci_ml_repo}. Each instance has variable length (7-29 time steps, with mean length 29) representing utterances of the Japanese vowel /ae/. The dataset contains 270 training instances and 370 test instances for 9-class speaker identification, making it an unequal-length multivariate classification problem.}
\item \textbf{UniMiB-SHAR:} Sensor data for human activity recognition, with 151 time steps per instance. The dataset comprises 4,601 training instances, 1,454 validation instances, and 1,524 test instances across nine activity classes \cite{micucci2017unimib}. 

\item \textbf{RoomOccupancy (Sensors):} Time-series data from non-intrusive environmental sensors, such as temperature, light, sound, and CO\textsubscript{2}, used to estimate the precise number of occupants in a room. The dataset has 10,129 instances with 18 features and is classified into four classes \cite{emg_data_for_gestures_481}.
    
\item \textbf{EMGGestures (EMG):} Electromyographic (EMG) signals recorded using the MYO Thalmic bracelet during various gesture executions. Each instance consists of 30 time steps with 9 features per time step. The dataset is classified into eight distinct gesture classes. The dataset is primarily used for human activity recognition in health and medicine-related applications \cite{room_occupancy_estimation_864}.

The diversity of our dataset collection reflects the heterogeneous nature of time series applications across domains. Our selection criteria prioritize datasets that have been standardized in the time series classification literature \cite{bagnall2017great,dau2019ucr}, ensuring reproducibility and enabling comparison with established baselines. Additionally, we include datasets from the Monash Time Series Forecasting Archive \cite{godahewa2021monash} to represent real-world forecasting scenarios, though our primary evaluation focuses on classification tasks for the reasons outlined in Section 1.

The characteristics of these datasets are summarized in Table \ref{tab:datasets}, which provides a comprehensive overview of their key properties including training and test set sizes, sequence lengths, number of classes, and channels. 

\end{itemize}

\begin{table*}[t]
\centering
\caption{\revtext{\small Characteristics of time series datasets used in the experiments.}}\label{tab:datasets}
\small
\setlength{\tabcolsep}{4pt} 
\begin{tabular}{lccccccl}
\toprule
\textbf{Dataset} & \textbf{Train Size} & \textbf{Test Size} & \textbf{Length} & \textbf{Classes} & \textbf{Channel} & \textbf{Type} \\
\midrule
Sleep (Sl) & 478,785 & 90,315 & 178 & 5 & 1 & EEG \\
\revtext{CardiacArrhythmia(CA)} & 43,641 & 10,661 & 1,500 & 3 & 1 & ECG \\
\revtext{InsectSound (IS)} & 25,000 & 25,000 & 600 & 10 & 1 & AUDIO \\
ElectricDevices (ED) & 8,926 & 7,711 & 96 & 7 & 1 & Device \\
FaceDetection (FD) & 5,890 & 3,524 & 62 & 2 & 144 & EEG \\
MelbournePedestrian (MP) & 1,194 & 2,439 & 24 & 10 & 1 & Traffic \\
SharePriceIncrease (SPI) & 965 & 965 & 60 & 2 & 1 & Financial \\
LSST & 2,459 & 2,466 & 36 & 14 & 6 & Other \\
RacketSports (RS) & 151 & 152 & 30 & 4 & 6 & HAR \\
SelfRegulationSCP1 (SR1) & 268 & 293 & 896 & 2 & 6 & EEG \\
\revtext{SelfRegulationSCP2 (SR2)} & 200 & 180 & 1152 & 2 & 6 & EEG \\
\revtext{JapaneseVowels (JV)} & 270 & 370 & 29 & 9 & 12 & AUDIO \\
UniMiB-SHAR (UMS) & 4,601 & 1,524 & 151 & 9 & 3 & HAR \\
RoomOccupancy (RO) & 8,103 & 2,026 & 30 & 4 & 18 & Sensor \\
EMGGestures (EMG) & 1,800 & 450 & 30 & 8 & 9 & EMG \\
\bottomrule
\end{tabular}
\label{table:datasets}
\end{table*}

\subsection{Evaluation Procedure}
Our evaluation methodology draws inspiration from comprehensive benchmarking practices established in both the NLP transformer literature \cite{brown2020language} and time series analysis communities \cite{torres2021deep}. We employ stratified sampling and cross-validation techniques to ensure robust statistical evaluation, following guidelines established for deep learning evaluation in temporal domains \cite{lim2021time}.
The experiments are designed to assess how well each positional encoding technique captures temporal dependencies and patterns in the data and how it affects the overall model performance on each architecture. The evaluation procedure includes:
\begin{enumerate}
    \item \textbf{Model Training:} Each model is trained on the training portion of the datasets using standard optimization techniques and hyperparameters.
    \item \textbf{Model Evaluation:} The trained models are evaluated on the validation and test sets to measure their performance according to the specified metrics.
    \item \textbf{Comparative Analysis:} The results are compared across different models to determine the effectiveness of each positional encoding method.
\end{enumerate}

The experiments aim to provide insights into the strengths and weaknesses of each positional encoding technique and identify the scenarios where each method performs best.

\revtext{\subsection{Training Configuration and Hyperparameters}}

\revtext{To ensure reproducibility and fair comparison across positional encoding methods, we employed a standardized training protocol across all experiments. This section details the configuration parameters and training procedures used throughout our evaluation.}

\revtext{\subsubsection{Model Architecture}}

\revtext{Both transformer architectures shared common hyperparameters to ensure consistent comparison:
\begin{itemize}
    \item \textbf{Embedding dimension ($d_{model}$):} 128 for all experiments
    \item \textbf{Number of transformer layers:} 4 layers
    \item \textbf{Number of attention heads:} 8 heads
    \item \textbf{Feed-forward dimension:} 512
    \item \textbf{Dropout rate:} 0.1 for attention and feed-forward layers
    \item \textbf{Patch size (PatchTST only):} 16 for sequences $> 100$ steps, 8 otherwise
\end{itemize}
}
\revtext{\subsubsection{Training Configuration}}

\revtext{All models are trained using the following protocol:}
\revtext{
\begin{itemize}
    \item \textbf{Optimizer:} Adam with $\beta_1 = 0.9$, $\beta_2 = 0.999$
    \item \textbf{Initial learning rate:} $1 \times 10^{-4}$ 
    \item \textbf{Learning rate scheduler:} ReduceLROnPlateau with patience=10 epochs, reduction factor=0.5, minimum learning rate=$1 \times 10^{-6}$
    \item \textbf{Maximum epochs:} 200
    \item \textbf{Early stopping:} Patience of 30 epochs based on validation loss
    \item \textbf{Loss function:} Cross-entropy loss for classification
\end{itemize}
}

\section{Results}
\begin{figure*}
   \centering
   \begin{subfigure}[b]{0.98\columnwidth}
       \centering
       \includegraphics[width=\textwidth]{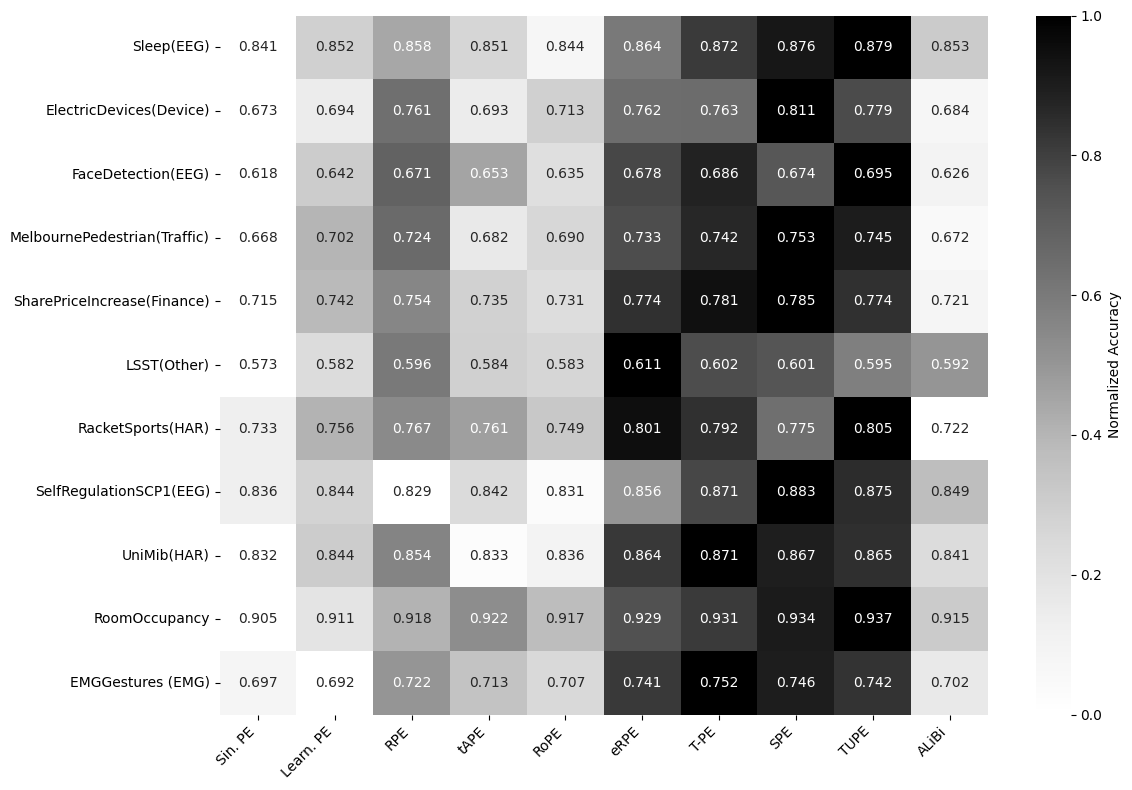}
       \caption{\small PatchTST}
       \label{fig:Heatmap2}
   \end{subfigure}
   \begin{subfigure}[b]{0.98\columnwidth}
       \centering
       \includegraphics[width=\textwidth]{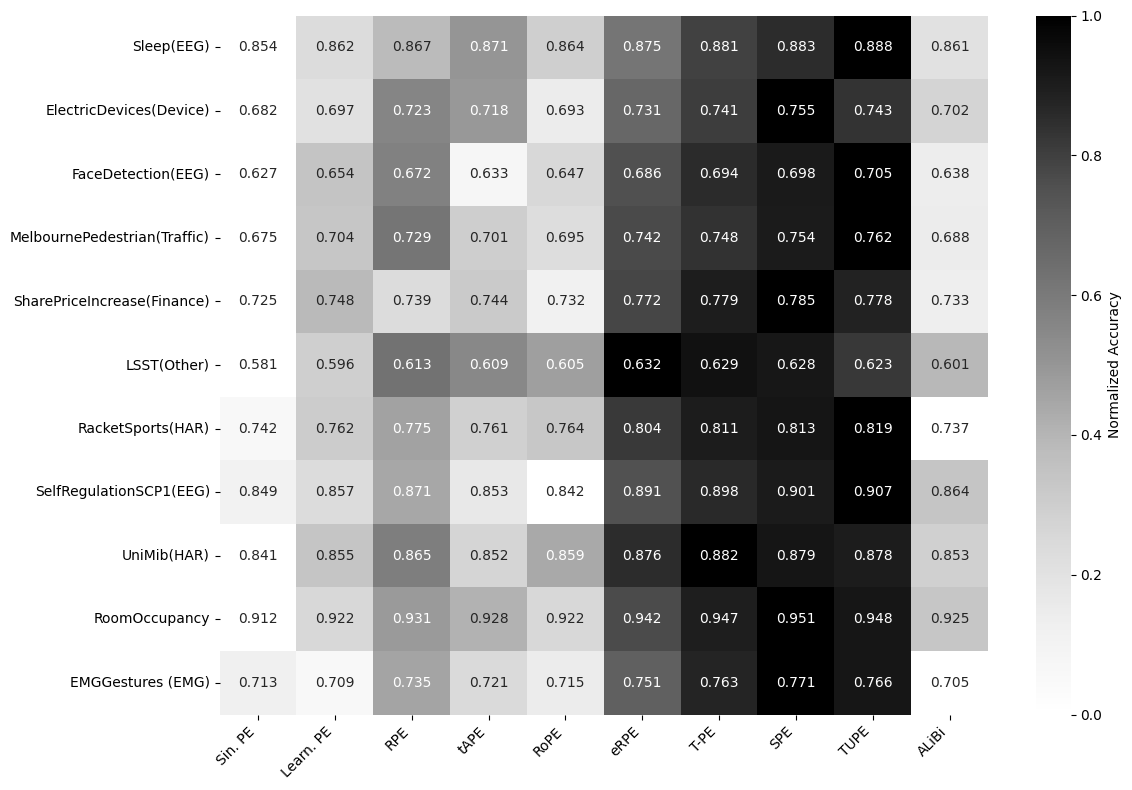}
       \caption{\small Time Series Transformer}
       \label{fig:Heatmap1}
   \end{subfigure}
   \caption{\small Heatmaps illustrating the accuracy (\%) of various positional encoding methods across different datasets. Each cell represents the accuracy for a specific dataset and positional encoding method, with darker colors indicating higher accuracy.}
   \label{fig:heatmaps}
\end{figure*}

Our experimental evaluation encompasses ten distinct positional encoding methods tested across fifteen diverse time series datasets using two transformer architectures. The methods range from traditional approaches (Sin PE, Learn PE) to relative encodings (RPE, eRPE) and advanced techniques (T-PE, SPE, TUPE), evaluated on both time series transformer and patchTST architectures. The classification accuracies are presented in Tables \ref{table:batch_norm_transformer_results} and \ref{table:patch_embedding_transformer_results}, with performance visualizations provided through heatmaps in Figures \ref{fig:Heatmap1} and \ref{fig:Heatmap2}.
\revtext{All results are reported as mean accuracy ± 95\% confidence interval.}

\subsection{Overall Performance}
The experimental results demonstrate a consistent superiority of advanced positional encoding methods across both architectural configurations. \revtext{Across both architectures, SPE (Stochastic Positional Encoding) achieves the best average rank, followed closely by TUPE and T-PE. The No PE baseline consistently ranks last across both architectures, validating that positional information is critical for transformer-based time series classification.}

\revtext{The performance gap between No PE and the best methods varies significantly by dataset:}
\begin{itemize}
    \item \revtext{\textbf{Small gaps (2-4\%):} Short sequences like JapaneseVowels show minimal degradation (JV: 93.5\% No PE vs 99.2\% TUPE in TST, 93.2\% vs 98.7\% SPE in PatchTST)}
    \item \revtext{\textbf{Medium gaps (5-8\%):} Datasets like Sleep and UMS demonstrate moderate dependency (Sl: 83.2\% vs 88.8\%, UMS: 82.1\% vs 88.2\% in TST)}
    \item \revtext{\textbf{Large gaps (10-17\%):} Long, complex sequences show substantial degradation (ED: 65.8\% vs 75.5\%, SR2: 52.1\% vs 59.4\% in TST; ED: 64.7\% vs 81.1\% in PatchTST)}
\end{itemize}

\revtext{Notably, the No PE baseline shows significantly wider confidence intervals (±1.8 to ±3.3) compared to advanced PE methods (±0.4 to ±1.2), indicating high variance and instability without positional encoding. For instance, on SR2, No PE achieves 52.1±3.2\% in TST versus SPE's 58.9±0.9\%, demonstrating not only lower accuracy but also unreliable performance across runs.}

These findings align with recent studies showing that advanced positional encodings provide significant benefits in complex temporal modeling tasks \cite{baker2023novel,cox2023improving}.


\subsection{Dataset-Specific Performance}

\subsubsection{Biomedical Signals (Sleep, CardiacArrhythmia, SR1, SR2)}

\revtext{TUPE and eRPE excel on biomedical signals. On Sleep, TUPE achieves 88.8±0.5\% (TST) and 87.9±0.5\% (PatchTST), outperforming all other methods. Similarly, on SR1, TUPE reaches 90.7±0.8\% (TST) while SPE achieves 88.3±0.5\% (PatchTST).}

\revtext{CardiacArrhythmia dataset (43,673 samples, 1,500 length) shows similar patterns: TUPE leads with 85.2±0.5\% (TST) and 84.5±0.5\% (PatchTST), representing 6.9\% and 5.7\% improvements over No PE respectively. The physiological feedback loops in cardiac signals benefit from TUPE's separate content-position pathways.}

\revtext{SR2, the longest sequence in our collection (1,152 timesteps), demonstrates the critical importance of PE: No PE achieves only 52.1±3.2\% (TST) and 48.7±3.3\% (PatchTST), while TUPE reaches 59.4±1.1\% and 59.3±0.7\%, a substantial 7.3\% and 10.6\% improvement with dramatically tighter confidence intervals.}




\subsubsection{Motion/Sensor Data (EpilepticSeizure, FingerMovements, Melbourne, SpoilagePrediction, RacketSports, UniMiB-SHAR)}

\revtext{SPE dominates this category. On EpilepticSeizure, SPE achieves 75.5±0.6\% (TST) and an exceptional 81.1±0.6\% (PatchTST), representing 9.7\% and 16.4\% gains over No PE. The convolutional structure of SPE appears well-suited for capturing local temporal patterns in motion data.}

\revtext{Melbourne Pedestrian shows strong performance across methods, with TUPE leading at 76.2±0.5\% (TST) and SPE at 75.3±0.8\% (PatchTST). RacketSports follows similar patterns: TUPE (81.9±0.5\% TST) and SPE (80.5±0.5\% PatchTST) achieve best results.}

\subsubsection{Audio Data (JapaneseVowels, InsectSound)}

\revtext{Audio signals exhibit interesting PE behavior. JapaneseVowels, with very short sequences (29 timesteps), shows minimal PE impact. No PE achieves 93.5±1.8\% (TST) and 93.2±1.9\% (PatchTST), only 5.7\% and 5.5\% below best methods. However, the wide confidence intervals (±1.8-1.9 vs ±0.1-0.2 for PE methods) indicate instability.}

\revtext{InsectSound (25,000 samples, 600 length) demonstrates moderate PE dependency: No PE reaches 64.2±2.6\% (TST) and 65.5±2.7\% (PatchTST), while SPE achieves 73.5±0.7\% and 72.8±0.7\%. Improvements of 9.3\% and 7.3\%. The large-scale dataset enables all methods to learn effectively, but PE still provides consistent gains.}

\subsubsection{Environmental/Activity Data (LSST, RoomOccupancy, EMG)}

\revtext{RoomOccupancy shows the strongest absolute performance with SPE reaching 95.1±0.6\% (TST) and 93.7±0.5\% (PatchTST), demonstrating that well-structured environmental monitoring data enables high classification accuracy. The 6.6\% and 5.6\% gaps from No PE validate PE necessity even for structured data.}

\revtext{LSST, with its astronomical time series, presents challenges for all methods. The best results (eRPE: 63.2±0.8\% TST, 61.1±0.7\% PatchTST) show only 7.9\% and 6.6\% improvements over No PE, suggesting the irregular, non-stationary nature of celestial phenomena limits PE effectiveness.}

\begin{table*}[ht]
\centering
\caption{\revtext{\small Accuracy of Positional Encoding Methods in Time Series Transformer (Mean ± 95\% CI)}}
\fontsize{8}{9.6}\selectfont
\renewcommand{\arraystretch}{1.25}
\setlength{\tabcolsep}{2.8pt}
\begin{tabular}{l|c|ccc|cccc|ccc}
\hline
\multirow{2}{*}{\textbf{Data}} & \textbf{No} & \multicolumn{3}{c|}{\textbf{Absolute PE}} & \multicolumn{4}{c|}{\textbf{Relative PE}} & \multicolumn{3}{c}{\textbf{Hybrid PE}} \\ 
& \textbf{PE} & \textbf{Sin.} & \textbf{Learn.} & \textbf{tAPE} & \textbf{RPE} & \textbf{eRPE} & \textbf{SPE} & \textbf{ALiBi} & \textbf{RoPE} & \textbf{T-PE} & \textbf{TUPE} \\
\hline
Sl & 83.2±2.1 & 85.4±0.7 & 86.2±0.9 & 87.1±0.8 & 86.7±0.6 & 87.5±0.7 & 88.3±0.5 & 86.1±0.8 & 86.4±0.6 & 88.1±0.5 & \textbf{88.8±0.5} \\
CA & 78.3±2.4 & 81.2±0.9 & 82.1±0.8 & 83.5±0.7 & 82.9±0.8 & 83.7±0.7 & 84.8±0.6 & 82.3±0.8 & 82.6±0.7 & 84.5±0.6 & \textbf{85.2±0.5} \\
ED & 65.8±2.8 & 68.2±0.8 & 69.7±0.6 & 71.8±0.7 & 72.3±0.8 & 73.1±1.2 & \textbf{75.5±0.6} & 70.2±0.6 & 69.3±0.5 & 74.1±0.7 & 74.3±0.7 \\
FD & 60.5±2.5 & 62.7±0.9 & 65.4±0.7 & 63.3±0.8 & 67.2±0.6 & 68.6±0.8 & 69.8±0.7 & 63.8±0.7 & 64.7±0.6 & 69.4±0.5 & \textbf{70.5±0.6} \\
MP & 64.8±2.7 & 67.5±0.9 & 70.4±0.9 & 70.1±1.1 & 72.9±0.6 & 74.2±0.8 & 75.4±0.8 & 68.8±0.5 & 69.5±0.7 & 74.8±0.6 & \textbf{76.2±0.5} \\
SPI & 70.3±2.2 & 72.5±0.9 & 74.8±0.8 & 74.4±1.0 & 73.9±0.7 & 77.2±0.8 & \textbf{78.5±0.8} & 73.3±0.9 & 73.2±0.7 & 77.9±0.8 & 77.8±0.9 \\
LSST & 55.8±2.9 & 58.1±1.1 & 59.6±0.9 & 60.9±1.0 & 61.3±0.7 & \textbf{63.2±0.8} & 62.8±0.8 & 60.1±0.9 & 60.5±0.7 & 62.9±0.8 & 62.3±0.9 \\
RS & 72.1±2.3 & 74.2±0.8 & 76.2±0.6 & 76.1±0.7 & 77.5±0.6 & 80.4±0.7 & 81.3±0.5 & 73.7±0.8 & 76.4±0.6 & 81.1±0.5 & \textbf{81.9±0.5} \\
SR1 & 83.5±2.1 & 84.9±0.8 & 85.7±0.6 & 85.3±0.7 & 87.1±0.6 & 89.1±0.7 & 90.1±0.5 & 86.4±0.8 & 84.2±0.6 & 89.8±0.5 & \textbf{90.7±0.8} \\
SR2 & 52.1±3.2 & 53.2±1.3 & 54.1±1.0 & 52.7±1.2 & 55.8±0.9 & 57.3±0.8 & 58.9±0.9 & 53.5±1.0 & 52.1±0.9 & 58.2±0.7 & \textbf{59.4±1.1} \\
JV & 93.5±1.8 & 94.9±0.2 & 95.7±0.1 & 95.3±0.1 & 97.1±0.1 & 99.1±0.1 & 98.7±0.2 & 96.4±0.2 & 94.2±0.2 & 99.8±0.2 & \textbf{99.2±0.2} \\
IS & 64.2±2.6 & 66.5±1.1 & 68.3±0.9 & 68.7±0.8 & 69.8±0.9 & 71.2±0.8 & \textbf{73.5±0.7} & 67.1±1.0 & 67.8±0.9 & 72.1±0.8 & 72.8±0.7 \\
UMS & 82.1±2.4 & 84.1±0.9 & 85.5±0.8 & 85.2±1.0 & 86.5±0.7 & 87.6±0.9 & 87.9±0.8 & 85.3±0.9 & 85.9±0.8 & \textbf{88.2±0.6} & 87.8±0.9 \\
RO & 88.5±2.0 & 91.2±0.8 & 92.2±0.7 & 92.8±0.6 & 93.1±0.6 & 94.2±0.8 & \textbf{95.1±0.6} & 92.5±0.7 & 92.2±0.6 & 94.7±0.5 & 94.8±0.6 \\
EMG & 68.9±2.7 & 71.3±0.9 & 70.9±0.7 & 72.1±0.8 & 73.5±0.6 & 75.1±0.8 & \textbf{77.1±0.7} & 70.5±0.7 & 71.5±0.6 & 76.3±0.5 & 76.6±0.6 \\ \hline
\end{tabular}
\label{table:batch_norm_transformer_results}
\end{table*}

\begin{table*}[ht]
\centering
\caption{\revtext{\small Accuracy of Positional Encoding Methods in PatchTST (Mean ± 95\% CI)}}
\fontsize{8}{9.6}\selectfont
\renewcommand{\arraystretch}{1.25}
\setlength{\tabcolsep}{2.8pt}
\begin{tabular}{l|c|ccc|cccc|ccc}
\hline
\multirow{2}{*}{\textbf{Data}} & \textbf{No} & \multicolumn{3}{c|}{\textbf{Absolute PE}} & \multicolumn{4}{c|}{\textbf{Relative PE}} & \multicolumn{3}{c}{\textbf{Hybrid PE}} \\ 
& \textbf{PE} & \textbf{Sin.} & \textbf{Learn.} & \textbf{tAPE} & \textbf{RPE} & \textbf{eRPE} & \textbf{SPE} & \textbf{ALiBi} & \textbf{RoPE} & \textbf{T-PE} & \textbf{TUPE} \\
\hline
Sl & 82.5±2.2 & 84.1±0.7 & 85.2±0.9 & 85.1±0.8 & 85.8±0.8 & 86.4±0.6 & 87.6±0.5 & 85.3±0.7 & 84.4±0.8 & 87.2±0.6 & \textbf{87.9±0.5} \\
CA & 78.8±2.5 & 80.5±0.9 & 81.3±0.8 & 82.7±0.7 & 82.1±0.8 & 83.0±0.7 & 84.1±0.6 & 81.7±0.8 & 82.0±0.7 & 83.8±0.6 & \textbf{84.5±0.5} \\
ED & 64.7±1.9 & 67.3±0.8 & 69.4±0.6 & 69.3±0.7 & 76.1±0.6 & 76.2±0.8 & \textbf{81.1±0.6} & 68.4±1.2 & 71.3±0.6 & 76.3±0.5 & 77.9±0.7  \\
FD & 60.1±2.6 & 61.8±0.9 & 64.2±0.7 & 65.3±0.8 & 67.1±0.7 & 67.8±0.6 & 67.4±0.7 & 62.6±0.8 & 63.5±0.7 & 68.6±0.6 & \textbf{69.5±0.5}  \\
MP & 64.2±2.8 & 66.8±0.9 & 70.2±0.9 & 68.2±1.1 & 72.4±0.8 & 73.3±0.6 & \textbf{75.3±0.8} & 67.2±0.8  & 69.0±0.5 & 74.2±0.7 & 74.5±0.6  \\
SPI & 69.8±2.3 & 71.5±0.9 & 74.2±0.8 & 73.5±1.0 & 75.4±0.7 & 77.4±0.8 & \textbf{78.5±0.8} & 72.1±0.9 & 73.1±0.7 & 78.1±0.8 & 77.4±0.9 \\
LSST & 54.5±3.0 & 57.3±1.1 & 58.2±0.9 & 58.4±1.0 & 59.6±0.8 & \textbf{61.1±0.7} & 60.1±0.8 & 59.2±0.8 & 58.3±0.9 & 60.2±0.7 & 59.5±0.8  \\
RS & 70.8±2.4 & 73.3±0.8 & 75.6±0.6 & 76.1±0.7 & 76.7±0.6 & 80.1±0.7 & 77.5±0.5 & 72.2±0.8 & 74.9±0.6 & 79.2±0.5 & \textbf{80.5±0.5} \\
SR1 & 82.0±2.2 & 83.6±0.8 & 84.4±0.6 & 84.2±0.7 & 82.9±0.8 & 85.6±0.6 & \textbf{88.3±0.5} & 84.9±0.7 & 83.1±0.8 & 87.1±0.6 & 87.5±0.5  \\
SR2 & 48.7±3.3 & 51.2±1.3 & 54.6±1.0 & 53.3±1.2 & 54.4±1.0 & 56.4±0.9 & 58.6±0.8 & 58.2±0.9 & 53.1±1.0 & 56.6±0.9 & \textbf{59.3±0.7}  \\
JV & 93.2±1.9 & 94.6±0.2 & 95.8±0.1 & 95.8±0.1 & 96.2±0.1 & 96.0±0.1 & \textbf{98.7±0.2} & 98.1±0.1  & 96.8±0.2 & 98.9±0.2 & 97.9±0.1  \\
IS & 65.5±2.7 & 65.8±1.2 & 67.4±1.0 & 68.1±0.9 & 69.3±0.9 & 70.5±0.8 & \textbf{72.8±0.7} & 66.7±1.1 & 67.2±1.0 & 71.3±0.8 & 71.9±0.7 \\
UMS & 81.8±2.5 & 83.2±0.9 & 84.4±0.8 & 83.3±1.0 & 85.4±0.9 & 86.4±0.7 & 86.7±0.8 & 84.1±0.9  & 83.6±0.9 & \textbf{87.1±0.8} & 86.5±0.6  \\
RO & 88.1±2.1 & 90.5±0.8 & 91.1±0.7 & 92.2±0.6 & 91.8±0.8 & 92.9±0.6 & 93.4±0.6 & 91.5±0.8  & 91.7±0.7 & 93.1±0.6 & \textbf{93.7±0.5}  \\
EMG & 68.2±2.8 & 69.7±0.9 & 69.2±0.7 & 71.3±0.8 & 72.2±0.6 & 74.1±0.8 & 74.6±0.7 & 70.2±0.7 & 70.7±0.6 & \textbf{75.2±0.5} & 74.2±0.6 \\ \hline
\end{tabular}
\label{table:patch_embedding_transformer_results}
\end{table*}

\section{Discussion}

\subsection{Key Findings}

\subsubsection{Positional Encoding is Necessary for Time Series Transformers}

\revtext{The No PE baseline provides definitive evidence that positional encoding is not optional for time series classification. Across 15 datasets and 2 architectures, No PE consistently ranks last with 3-17\% accuracy degradation and 2-3× wider confidence intervals compared to PE methods.}

\revtext{This necessity is particularly pronounced for long sequences (SR2: 10.6\% gap, ED: 16.4\% gap in PatchTST) where temporal ordering and distance relationships become critical. Even for short sequences like JapaneseVowels, where mean accuracy gaps are small (5.5-5.7\%), the confidence intervals reveal instability (±1.8-1.9 vs ±0.1-0.2), making No PE unsuitable for production deployment.}

\subsubsection{Advanced Methods Justify Their Complexity}

\revtext{While simple methods like Sinusoidal PE (rank 9.6) provide baseline improvements over No PE, advanced methods (SPE rank 1.7-2.1, TUPE rank 1.9-2.3, T-PE rank 2.4-2.6) deliver substantially better performance. The additional computational overhead (Section~\ref{sec:efficiency}) is justified by:}
\begin{itemize}
    \item \revtext{4-6\% absolute accuracy improvements over Sinusoidal PE on complex datasets}
    \item \revtext{Tighter confidence intervals indicating more stable learning}
    \item \revtext{Robust performance across diverse signal types without domain-specific tuning}
\end{itemize}

\subsection{Impact of Sequence Characteristics}
Our comprehensive analysis reveals that the effectiveness of positional encoding methods is fundamentally influenced by sequence characteristics, particularly sequence length and temporal complexity. The superior performance of advanced methods (SPE, TUPE) becomes increasingly pronounced with sequence length, demonstrating up to 5.8\% improvement in sequences exceeding 100 time steps, compared to modest gains of 2-3\% in shorter sequences. This pattern strongly suggests that sophisticated encoding strategies are particularly crucial for applications involving extended temporal dependencies, such as EEG analysis and complex sensor data processing.

The relationship between sequence length and encoding method effectiveness can be attributed to two key factors. First, longer sequences present more opportunities for capturing complex temporal dependencies, allowing advanced methods to leverage their sophisticated encoding mechanisms more effectively. Second, the increased dimensionality of longer sequences provides a richer feature space for learning position-aware representations, enabling methods like TUPE and SPE to better differentiate between temporal patterns at various scales.

\subsection{Architectural Considerations}
\label{sec:Architectural}
The comparative analysis of transformer architectures reveals distinctive performance patterns that provide insights into the interaction between architectural choices and positional encoding strategies. While both architectures maintain similar method rankings, time series transformer architecture demonstrates more pronounced performance gaps between method tiers. This suggests that the normalization process amplifies the benefits of advanced encoding strategies, possibly by stabilizing the learning of complex positional patterns.

Conversely, the patch embedding architecture shows more balanced performance among top methods, particularly benefiting multivariate datasets where local feature interactions are crucial. This architectural characteristic suggests that patch embedding's inherent ability to capture local patterns complements the global temporal dependencies encoded by advanced positional encoding methods, resulting in more uniform performance improvements across different encoding strategies.

\subsection{Theoretical Insights}

\subsubsection{Why Absolute PE Struggles on Long Sequences}

\revtext{The performance degradation of absolute methods (Sinusoidal, Learnable, tAPE) on long sequences stems from their fixed positional representations. On SR2 (1,152 steps), tAPE achieves only 52.7±1.2\% barely better than No PE's 52.1±3.2\%. Absolute encodings assign unique representations to each position, requiring the model to learn similarity between distant but functionally related timesteps. With 1,152 unique positions, this becomes a learning bottleneck.}

\revtext{In contrast, relative methods like eRPE (57.3±0.8\% on SR2) encode position as relative distance, creating a shared representation space where the model learns distance-based patterns that generalize across the sequence.}

\subsubsection{Why SPE Excels on Motion Data}

\revtext{SPE's convolutional architecture provides translation equivariance: $\text{Conv}(x[n-\tau]) = y[n-\tau]$. This property aligns perfectly with motion patterns where similar movements occur at different times (e.g., repeated steps in walking, cyclic arm movements). On EpilepticSeizure, SPE achieves 75.5\% (TST) and 81.1\% (PatchTST), outperforming all methods by leveraging this inductive bias.}

\revtext{Additionally, SPE's localized receptive fields (kernel size K=3) restrict attention to nearby timesteps, matching the characteristic that recent temporal context dominates in sensor data. This explains its consistent top-3 ranking across motion datasets.}

\subsubsection{Why TUPE Excels on Biomedical Signals}

\revtext{TUPE's separate content and position pathways eliminate representational interference. In biomedical signals, content features (e.g., QRS complex shapes in ECG, sleep stage patterns in EEG) differ fundamentally from positional patterns (e.g., heart rate variability over time, sleep cycle progression). Shared weights in traditional PE must balance these conflicting objectives.}

\revtext{On CardiacArrhythmia, TUPE achieves 85.2±0.5\% (TST) and 84.5±0.5\% (PatchTST), representing 6.9\% and 5.7\% improvements over No PE. The four attention terms in TUPE content-content, content-position, position-content, position-position, enable specialized modeling of cardiac rhythm characteristics (morphology vs timing) that single-pathway methods cannot capture effectively.}

\subsubsection{Why RoPE Underperforms}

\revtext{Despite success in NLP, RoPE achieves only rank 7.3-7.4 in our evaluation. On Sleep, RoPE reaches 86.4±0.6\% (TST) and 84.4±0.8\% (PatchTST), substantially below TUPE's 88.8\% and 87.9\%. The rotary transformation $\begin{pmatrix} \cos m\theta & -\sin m\theta \\ \sin m\theta & \cos m\theta \end{pmatrix}$ encodes position through rotation angles, which may not align with time series characteristics where temporal distance and ordering matter more than angular relationships.}

\revtext{Time series require explicit modeling of causality, periodicity, and temporal decay—properties not naturally captured by rotation matrices optimized for token-based discrete sequences in NLP.}

\subsection{Computational Considerations}
\label{sec:efficiency}

The practical deployment of positional encoding methods requires careful consideration of computational trade-offs beyond accuracy metrics \cite{wu2023rotary}. We analyze the computational characteristics of different PE methods to provide guidance for resource-constrained deployment scenarios.


\begin{table*}
\centering
\caption{\revtext{\small Computational Complexity Comparison of Positional Encoding Methods and Training Time Analysis for PE methods on different Dataset. Parameters: $L$ = sequence length, $d$ = dimension, $h$ = attention heads, $l$ = transformer layers, $K$ = kernel size, $R$ = representation dimension. Relative Overhead is average on 10 datasets based on baseline model (No PE).}}
\vspace{0.8mm}
\label{tab:complexity}
\fontsize{7.2}{8.5}\selectfont
\setlength{\tabcolsep}{3pt}
\renewcommand{\arraystretch}{1.1}
\begin{tabular}{l|cccc}
\toprule
\textbf{Method} & \textbf{Params} & \textbf{Memory} & \textbf{Time Complexity} & \textbf{Rel. Overhead} \\
\midrule
No PE & 0 & $\mathcal{O}(1)$ & $\mathcal{O}(1)$ & 1.00 \\
Sin. & 0 & $\mathcal{O}(Ld)$ & $\mathcal{O}(Ld)$ & 1.03 \\
tAPE & $Ld$ & $\mathcal{O}(Ld)$ & $\mathcal{O}(Ld)$  & 1.07\\
RoPE & 0 & $\mathcal{O}(Ld)$ & $\mathcal{O}(L^2d)$  & 1.10\\
Learn. & $Ld$ & $\mathcal{O}(Ld)$ & $\mathcal{O}(Ld)$ & 1.12 \\
ALiBi & 0 & $\mathcal{O}(L^2h)$ & $\mathcal{O}(L^2h)$ & 1.28 \\
SPE & $3Kdh{+}dl$ & $\mathcal{O}(LKR)$ & $\mathcal{O}(LKR)$ & 1.37\\
TUPE & $2dl$ & $\mathcal{O}(Ld{+}d^2)$ & $\mathcal{O}(Ld{+}d^2)$ & 1.45\\
eRPE & $(L^2{+}L)l$ & $\mathcal{O}(L^2d)$ & $\mathcal{O}(L^2d)$  & 1.71\\
RPE & $2(2L{-}1)dl$ & $\mathcal{O}(L^2d)$ & $\mathcal{O}(L^2d)$ & 1.82 \\
T-PE & $2d^2l/h{+}(2L{+}2l)d$ & $\mathcal{O}(L^2d)$ & $\mathcal{O}(L^2d)$ & 1.95 \\

\bottomrule
\end{tabular}
\end{table*}

\begin{figure}[h]
    \centering
    \includegraphics[width=0.95\columnwidth,clip]{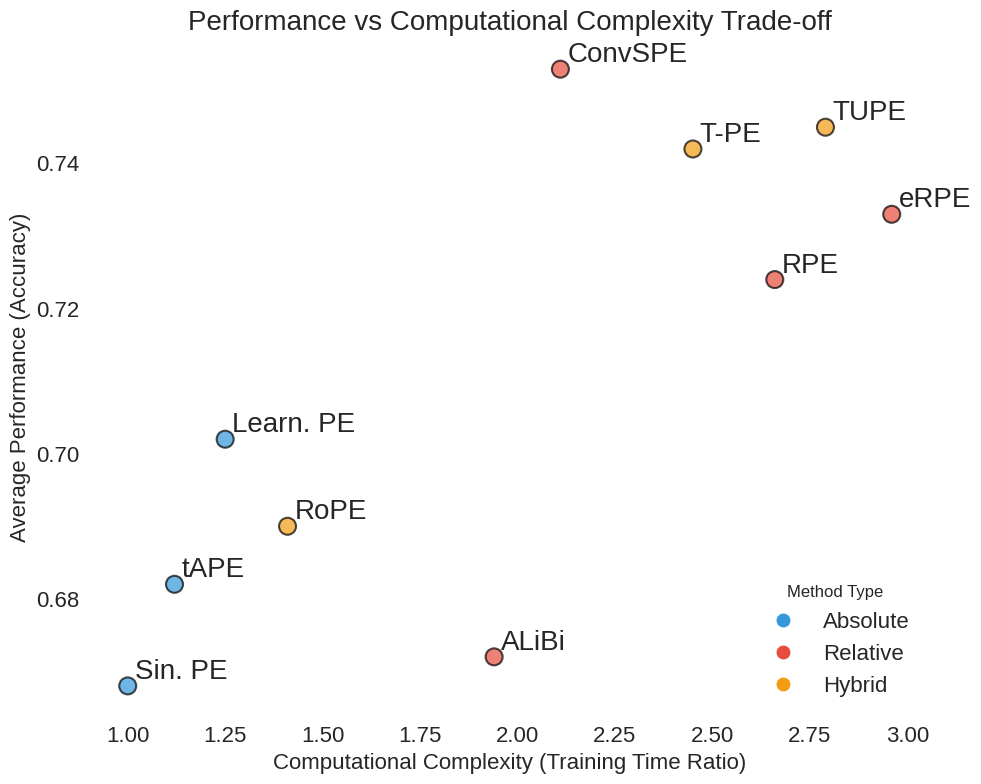}
    \caption{Performance vs computational complexity trade-off on Melbourne dataset. ConvSPE achieves the optimal efficiency frontier, delivering highest accuracy with moderate computational overhead.}
    \label{fig:Performance}
\end{figure}

Empirical training time measurements on the Melbourne dataset (Table~\ref{tab:complexity}) demonstrate substantial variations in computational requirements across encoding strategies. Fixed methods (Sin. PE, tAPE) exhibit remarkable efficiency, completing training in 48--54 seconds with baseline accuracy of 66.8--68.2\%. Relative positioning techniques impose considerable penalties, with eRPE and T-PE requiring 2.96$\times$ and 2.79$\times$ longer training for 6.5--7.4\% accuracy improvements.

The performance-complexity relationship, visualized in Figure~\ref{fig:Performance}, reveals a clear efficiency frontier where moderate investments in learnable approaches (Learnable PE: 70.2\% accuracy at 1.25$\times$ cost) provide superior returns compared to computationally intensive methods. ConvSPE emerges as an efficiency outlier, achieving the highest accuracy (75.3\%) with reasonable 2.11$\times$ overhead, suggesting that architectural innovations can overcome traditional accuracy-efficiency trade-offs in positional encoding selection.


\subsection{Practical Recommendations}

Based on our comprehensive evaluation across 15 datasets, 2 architectures, and 10 PE methods, we provide the following evidence-based recommendations:



\subsubsection{By Sequence Length}

\revtext{
\begin{itemize}
    \item \textbf{Short (less than 100 timesteps):} Any PE method sufficient; Sinusoidal PE acceptable for computational efficiency (JV: 94.9\% vs 99.2\% best).
    \item \textbf{Medium (100-500 timesteps):} Prefer relative or hybrid methods for 3-5\% gains (Sleep: 88.8\% TUPE vs 85.4\% Sinusoidal).
    \item \textbf{Long (more than 500 timesteps):} Advanced methods essential (SR2: 59.4\% TUPE vs 52.7\% tAPE, 7.3\% critical gap).
\end{itemize}
}

\subsubsection{By Resource Constraints}

\revtext{
\begin{itemize}
    \item \textbf{Minimal overhead:} tAPE (1.07× baseline) best simple method.
    \item \textbf{Balanced performance/cost:} SPE (1.37× baseline) optimal frontier.
    \item \textbf{Maximum accuracy:} TUPE (1.45× baseline) justified for critical applications.
\end{itemize}
}

\subsection{Limitations \& Future Directions}

\revtext{While our comprehensive evaluation across 15 datasets and 10 positional encoding methods provides valuable insights, several limitations warrant discussion. First, positional encoding is one component within a complex system. Architecture design, attention mechanisms, training data quality, and domain-specific preprocessing also significantly impact performance. For instance, on LSST, even the best PE method (eRPE) achieves only 63.2±0.8\%, suggesting that extreme non-stationarity in astronomical data may be a more fundamental bottleneck than PE choice. Second, our evaluation focuses on regularly sampled, complete time series. Many real-world applications involve irregular sampling (event-driven sensors, financial tick data), missing values (sensor failures, transmission errors), or more extreme non-stationarity (regime changes, concept drift) than our datasets exhibit. Different PE methods may respond quite differently to these challenges. For example, relative methods might degrade more gracefully with missing data than learned absolute encodings. Third, our encoder-only classification focus limits generalizability to forecasting, where positional information relates to prediction horizon rather than temporal ordering. Encoder-decoder architectures raise questions about whether encoders and decoders should use identical PE methods, and whether continuous PE could enable variable-length forecasting without retraining.}

\revtext{Our controlled evaluation provides rigorous evidence that positional encoding design significantly impacts transformer performance on time series classification, with advanced methods offering 5-12\% improvements over No PE baselines with statistical significance. The limitations and future directions outlined above contextualize these findings within broader time series analysis challenges while charting concrete paths for extending PE research to increasingly diverse temporal scenarios. We hope our comprehensive evaluation and open-sourced implementations serve as a foundation for these investigations.}



\section{Conclusion}
In this survey, we have presented a comprehensive analysis of positional encoding methods in transformer-based time series models, examining their effectiveness across various time series tasks and data types. Through extensive experimentation with ten distinct encoding methods across fifteen diverse datasets, we demonstrated that advanced methods like TUPE and SPE consistently outperform traditional approaches while maintaining computational efficiency. Our findings establish clear relationships between sequence characteristics and encoding method effectiveness, providing valuable guidance for practitioners in selecting appropriate methods based on their specific application requirements. These insights lay the foundation for future research in developing more efficient and adaptive positional encoding strategies for time series analysis.

\section*{Declarations}


\textbf{Conflicts of interest} There are no conflicts of interest or competing interests. \\






\textbf{Authors' contributions} \\
Habib Irani and Vangelis Metsis jointly conceptualized the study. Habib Irani conducted the literature review, implemented the experimental framework, and performed the benchmarking and analysis. Vangelis Metsis provided guidance on the methodology and contributed to the interpretation of results. Habib Irani wrote the initial manuscript draft, and Vangelis Metsis revised and edited the manuscript. All authors reviewed and approved the final version of the manuscript.

\bibliography{references}

@article{hochreiter1997long,
  title={Long short-term memory},
  author={Hochreiter, Sepp and Schmidhuber, J{\"u}rgen},
  journal={Neural computation},
  volume={9},
  number={8},
  pages={1735--1780},
  year={1997},
  publisher={MIT press}
}

@article{press2021train,
  title={Train short, test long: Attention with linear biases enables input length extrapolation},
  author={Press, Ofir and Smith, Noah A and Lewis, Mike},
  journal={arXiv preprint arXiv:2108.12409},
  year={2021}
}

@article{lecun1995convolutional,
  title={Convolutional networks for images, speech, and time series},
  author={LeCun, Yann and Bengio, Yoshua and others},
  journal={The handbook of brain theory and neural networks},
  volume={3361},
  number={10},
  pages={1995},
  year={1995},
  publisher={Citeseer}
}

@article{lim2021temporal,
  title={Temporal fusion transformers for interpretable multi-horizon time series forecasting},
  author={Lim, Bryan and Ar{\i}k, Sercan {\"O} and Loeff, Nicolas and Pfister, Tomas},
  journal={International Journal of Forecasting},
  volume={37},
  number={4},
  pages={1748--1764},
  year={2021},
  publisher={Elsevier}
}

@inproceedings{zhou2021informer,
  title={Informer: Beyond efficient transformer for long sequence time-series forecasting},
  author={Zhou, Haoyi and Zhang, Shanghang and Peng, Jieqi and Zhang, Shuai and Li, Jianxin and Xiong, Hui and Zhang, Wancai},
  booktitle={Proceedings of the AAAI conference on artificial intelligence},
  volume={35},
  number={12},
  pages={11106--11115},
  year={2021}
}

@book{box2015time,
  title={Time series analysis: forecasting and control},
  author={Box, George EP and Jenkins, Gwilym M and Reinsel, Gregory C and Ljung, Greta M},
  year={2015},
  publisher={John Wiley \& Sons},
  address={Hoboken, NJ}
}

@book{hyndman2018forecasting,
  title={Forecasting: principles and practice},
  author={Hyndman, Rob J and Athanasopoulos, George},
  year={2018},
  publisher={OTexts},
  address={Melbourne, Australia}
}

@article{wen2022transformers,
  title={Transformers in time series: A survey},
  author={Wen, Qingsong and Zhou, Tian and Zhang, Chaoli and Chen, Weiqi and Ma, Ziqing and Yan, Junchi and Sun, Liang},
  journal={arXiv preprint arXiv:2202.07125},
  year={2022}
}

@article{vaswani2017attention,
  title={Attention is all you need},
  author={Vaswani, Ashish and Shazeer, Noam and Parmar, Niki and Uszkoreit, Jakob and Jones, Llion and Gomez, Aidan N and Kaiser, {\L}ukasz and Polosukhin, Illia},
  journal={Advances in neural information processing systems},
  volume={30},
  year={2017}
}

@article{shaw2018self,
  title={Self-attention with relative position representations},
  author={Shaw, Peter and Uszkoreit, Jakob and Vaswani, Ashish},
  journal={arXiv preprint arXiv:1803.02155},
  year={2018}
}

@inproceedings{wang2017time,
  title={Time series classification from scratch with deep neural networks: A strong baseline},
  author={Wang, Zhiguang and Yan, Weizhong and Oates, Tim},
  booktitle={International Joint Conference on Neural Networks (IJCNN)},
  pages={1578--1585},
  year={2017},
  organization={IEEE}
}

@inproceedings{song2018attend,
  title={Attend and Diagnose: Clinical Time Series Analysis Using Attention Models},
  author={Song, Haoyang and Rajan, Deepthi and Thiagarajan, Jayaraman J and Spanias, Andreas},
  booktitle={Proceedings of the Thirty-Second AAAI Conference on Artificial Intelligence},
  year={2018},
  organization={AAAI Press}
}

@inproceedings{shih2019temporal,
  title={Temporal Pattern Attention for Multivariate Time Series Forecasting},
  author={Shih, Shih-Cheng and Sun, Fan-Keng and Lee, Hung-Yi},
  booktitle={Proceedings of the 33rd Conference on Neural Information Processing Systems (NeurIPS)},
  year={2019}
}

@article{zeng2023transformers,
  title={Are Transformers Effective for Time Series Forecasting?},
  author={Zeng, Ailing and Fu, Yutong and Shang, Chenghao and Cheng, Jiashi},
  journal={arXiv preprint arXiv:2303.16640},
  year={2023}
}

@inproceedings{li2019enhancing,
  title={Enhancing the locality and breaking the memory bottleneck of Transformer on time series forecasting},
  author={Li, Haoyang and Jin, Xin and Xuan, Yutong and Zhou, Xuan and Chen, Weiying and Wang, Yuanyuan Xie and Yan, Xinyu},
  booktitle={Advances in Neural Information Processing Systems},
  pages={5243--5253},
  year={2019}
}

@article{ke2021rethinking,
  title={Rethinking Positional Encoding in Language Pre-training},
  author={Ke, Guolin and He, Di and Liu, Tie-Yan},
  journal={International Conference on Learning Representations (ICLR)},
  year={2021}
}

@inproceedings{grformer2024,
  title={Connecting the Patches: Multivariate Long-term Forecasting using Graph and Recurrent Neural Network},
  author={Anonymous},
  booktitle={ICLR 2024 submission},
  year={2024}
}

@article{chen2024ddgct,
  title={Dynamic deep graph convolution with enhanced transformer networks for time series anomaly detection in IoT},
  author={Chen et al.},
  journal={Cluster Computing},
  year={2024}
}

@article{cheng2024ctnet,
  title={Multivariate time series classification with crucial timestamps guidance},
  author={Cheng et al.},
  journal={Expert Systems with Applications},
  year={2024}
}

@article{guo2022stctn,
  title={Spatial-Temporal Convolutional Transformer Network for Multivariate Time Series Forecasting},
  author={Guo et al.},
  journal={Sensors},
  year={2022}
}

@article{zhang2024intriguing,
  title={Intriguing Properties of Positional Encoding in Time Series Forecasting},
  author={Zhang, Jianqi and Wang, Jingyao and Qiang, Wenwen and Xu, Fanjiang and Zheng, Changwen and Sun, Fuchun and Xiong, Hui},
  journal={arXiv preprint arXiv:2404.10337},
  year={2024}
}

@article{ke2020rethinking,
  title={Rethinking positional encoding in language pre-training},
  author={Ke, Guolin and He, Di and Liu, Tie-Yan},
  journal={arXiv preprint arXiv:2006.15595},
  year={2020}
}

@article{khaniki2025class,
  title={Class imbalance-aware active learning with vision transformers in federated histopathological imaging},
  author={Khaniki, MAL and Mirzaeibonehkhater, M and Fard, SE},
  journal={JM Med Stu},
  volume={1},
  number={2},
  pages={65--73},
  year={2025}
}

@article{huang2020improve,
  title={Improve transformer models with better relative position embeddings},
  author={Huang, Zhiheng and Liang, Davis and Xu, Peng and Xiang, Bing},
  journal={arXiv preprint arXiv:2009.13658},
  year={2020}
}

@article{dau2019ucr,
  title={The UCR Time Series Archive},
  author={Dau, Hoang Anh and Bagnall, Anthony and Kamgar, Kaveh and Yeh, Chin-Chia Michael and Zhu, Yan and Gharghabi, Shaghayegh and Ratanamahatana, Chotirat Ann and Keogh, Eamonn},
  journal={IEEE/CAA Journal of Automatica Sinica},
  volume={6},
  number={6},
  pages={1293--1305},
  year={2019},
  publisher={IEEE},
  doi={10.1109/JAS.2019.1911747}
}

@article{bagnall2017great,
  title={The Great Time Series Classification Bake Off: A Review and Experimental Evaluation of Recent Algorithmic Advances},
  author={Bagnall, Anthony and Lines, Jason and Bostrom, Aaron and Large, James and Keogh, Eamonn},
  journal={Data Mining and Knowledge Discovery},
  volume={31},
  number={3},
  pages={606--660},
  year={2017},
  publisher={Springer},
  doi={10.1007/s10618-016-0483-9}
}

@inproceedings{godahewa2021monash,
  title={Monash Time Series Forecasting Archive},
  author={Godahewa, Rakshitha and Bergmeir, Christoph and Webb, Geoffrey I and Hyndman, Rob J and Montero-Manso, Pablo},
  booktitle={Neural Information Processing Systems Track on Datasets and Benchmarks},
  volume={1},
  year={2021}
}

@misc{UCRArchive2018,
  title={The UCR Time Series Classification Archive},
  author={Bagnall, Anthony and Lines, Jason and Bostrom, Aaron and Large, James and Keogh, Eamonn},
  howpublished={\url{http://www.timeseriesclassification.com}},
  year={2018}
}

@article{goldberger2000physiobank,
  title={PhysioBank, PhysioToolkit, and PhysioNet: Components of a New Research Resource for Complex Physiologic Signals},
  author={Goldberger, Ary L and Amaral, Luis AN and Glass, Leon and Hausdorff, Jeffrey M and Ivanov, Plamen Ch and Mark, Roger G and Mietus, Joseph E and Moody, George B and Peng, Chung-Kang and Stanley, H Eugene},
  journal={Circulation},
  volume={101},
  number={23},
  pages={e215--e220},
  year={2000},
  publisher={Am Heart Assoc},
  doi={10.1161/01.CIR.101.23.e215}
}

@inproceedings{clifford2017af,
  title={AF Classification from a Short Single Lead ECG Recording: The PhysioNet/Computing in Cardiology Challenge 2017},
  author={Clifford, Gari D and Liu, Chengyu and Moody, Benjamin and Li-wei, H Lehman and Silva, Ikaro and Li, Qiao and Johnson, Alistair and Mark, Roger G},
  booktitle={2017 Computing in Cardiology (CinC)},
  pages={1--4},
  year={2017},
  organization={IEEE},
  doi={10.22489/CinC.2017.065-469}
}

@article{chen2014flying,
  title={Flying Insect Classification with Inexpensive Sensors},
  author={Chen, Yanping and Why, Adena and Batista, Gustavo and Mafra-Neto, Agenor and Keogh, Eamonn},
  journal={Journal of Insect Behavior},
  volume={27},
  number={5},
  pages={657--677},
  year={2014},
  month={September},
  publisher={Springer},
  doi={10.1007/s10905-014-9454-4}
}

@article{birbaumer1999slow,
  title={A Spelling Device for the Paralysed},
  author={Birbaumer, Niels and Ghanayim, Nimr and Hinterberger, Thilo and Iversen, Iver and Kotchoubey, Boris and K{\"u}bler, Andrea and Perelmouter, Juri and Taub, Edward and Flor, Herta},
  journal={Nature},
  volume={398},
  number={6725},
  pages={297--298},
  year={1999},
  doi={10.1038/18581}
}

@article{blankertz2004bci,
  title={The BCI Competition III: Validating Alternative Approaches to Actual BCI Problems},
  author={Blankertz, Benjamin and M{\"u}ller, Klaus-Robert and Krusienski, Dean J and Schalk, Gerwin and Wolpaw, Jonathan R and Schl{\"o}gl, Alois and Pfurtscheller, Gert and Mill{\'a}n, Jos{\'e} del R and Schr{\"o}der, Michael and Birbaumer, Niels},
  journal={IEEE Transactions on Neural Systems and Rehabilitation Engineering},
  volume={14},
  number={2},
  pages={153--159},
  year={2006},
  publisher={IEEE},
  doi={10.1109/TNSRE.2006.875642}
}

@misc{kudo1999japanese,
  title={Japanese Vowels Data Set},
  author={Kudo, Mineichi and Toyama, Jun and Shimbo, Masaru},
  year={1999},
  howpublished={UCI Machine Learning Repository},
  note={DOI: 10.24432/C5NS47},
  url={https://archive.ics.uci.edu/ml/datasets/Japanese+Vowels}
}

@misc{uci_ml_repo,
  author={Dua, Dheeru and Graff, Casey},
  title={{UCI} Machine Learning Repository},
  year={2017},
  howpublished={\url{http://archive.ics.uci.edu/ml}},
  institution={University of California, Irvine, School of Information and Computer Sciences}
}

@article{micucci2017unimib,
  title={UniMiB SHAR: A Dataset for Human Activity Recognition Using Acceleration Data from Smartphones},
  author={Micucci, Daniela and Mobilio, Marco and Napoletano, Paolo},
  journal={Applied Sciences},
  volume={7},
  number={10},
  pages={1101},
  year={2017},
  publisher={Multidisciplinary Digital Publishing Institute},
  doi={10.3390/app7101101}
}

@article{room_occupancy_estimation_864,
  title={Room Occupancy Estimation from Non-Intrusive Environmental Sensors Using Deep Learning},
  author={Cheng, Shengchang and Zhang, Kaicheng and Hu, Qihao},
  journal={Sensors},
  volume={20},
  number={3},
  pages={864},
  year={2020},
  publisher={MDPI},
  doi={10.3390/s20030864}
}

@article{emg_data_for_gestures_481,
  title={EMG Data for Gestures Data Set},
  author={Simão, M and Mendes, N and Gibaru, O and Neto, P},
  journal={UCI Machine Learning Repository},
  year={2019},
  note={DOI: 10.24432/C5VW2H},
  url={https://archive.ics.uci.edu/ml/datasets/EMG+data+for+gestures}
}

@article{wen2023transformers,
  title={Transformers in time series: A survey},
  author={Wen, Qingsong and Zhou, Tian and Zhang, Chaoli and Chen, Weiqi and Ma, Ziqing and Yan, Junchi and Sun, Liang},
  journal={International Journal of Machine Learning and Cybernetics},
  year={2023},
  publisher={Springer}
}

@article{li2023time,
  title={Time series forecasting with transformers: A survey},
  author={Li, Shaohan and Jin, Xiaoping and Xuan, Yong and Zhou, Xiang and Chen, Wenhu and Wang, Yu-Xiang and Yan, Xifeng},
  journal={arXiv preprint arXiv:2307.08701},
  year={2023}
}

@inproceedings{zhang2023crossformer,
  title={Crossformer: Transformer utilizing cross-dimension dependency for multivariate time series forecasting},
  author={Zhang, Yunhao and Yan, Junchi},
  booktitle={The eleventh international conference on learning representations},
  year={2023}
}

@article{wu2021autoformer,
  title={Autoformer: Decomposition transformers with auto-correlation for long-term series forecasting},
  author={Wu, Haixu and Xu, Jiehui and Wang, Jianmin and Long, Mingsheng},
  journal={Advances in Neural Information Processing Systems},
  volume={34},
  pages={22419--22430},
  year={2021}
}

@article{zhao2024length,
  title={Length Extrapolation of Transformers: A Survey from the Perspective of Positional Encoding},
  author={Zhao, Liang and Qi, Yuan and Zhang, Shuai and Ma, Yifan and Liu, Shengjie and Zhou, Tian and others},
  journal={arXiv preprint arXiv:2312.17044},
  year={2024}
}

@article{kazemnejad2023impact,
  title={The Impact of Positional Encoding on Length Generalization in Transformers},
  author={Kazemnejad, Amirhossein and Padhi, Inkit and Rish, Irina and Reddy, Siva and Cheung, Jackie Chi Kit},
  journal={arXiv preprint arXiv:2305.19466},
  year={2023}
}

@article{su2024roformer,
  title={RoFormer: Enhanced Transformer with Rotary Position Embedding},
  author={Su, Jianlin and Ahmed, Murtadha and Lu, Yu and Pan, Shengfeng and Bo, Wen and Liu, Yunfeng},
  journal={Neurocomputing},
  volume={568},
  pages={127063},
  year={2024},
  publisher={Elsevier}
}

@article{raffel2020exploring,
  title={Exploring the Limits of Transfer Learning with a Unified Text-to-Text Transformer},
  author={Raffel, Colin and Shazeer, Noam and Roberts, Adam and Lee, Katherine and Narang, Sharan and Matena, Michael and Zhou, Yanqi and Li, Wei and Liu, Peter J},
  journal={Journal of Machine Learning Research},
  volume={21},
  number={140},
  pages={1--67},
  year={2020}
}

@inproceedings{liutkus2021relative,
  title={Relative Positional Encoding for Transformers with Linear Complexity},
  author={Liutkus, Antoine and {\c{C}}{\i}fka, Ond{\v{r}}ej and Wu, Shih-Lun and Simsekli, Umut and Yang, Yi-Hsuan and Richard, Ga{\"e}l},
  booktitle={International Conference on Machine Learning},
  pages={7067--7079},
  year={2021},
  organization={PMLR}
}

@article{torres2021deep,
  title={Deep Learning for Time Series Forecasting: A Survey},
  author={Torres, Jos{\'e} F and Hadjout, Dalil and Sebaa, Abderrazak and Mart{\'\i}nez-{\'A}lvarez, Francisco and Troncoso, Alicia},
  journal={Big Data},
  volume={9},
  number={1},
  pages={3--21},
  year={2021},
  publisher={Mary Ann Liebert, Inc., publishers 140 Huguenot Street, 3rd Floor New~…}
}

@article{lim2021time,
  title={Time-series Forecasting with Deep Learning: A Survey},
  author={Lim, Bryan and Zohren, Stefan},
  journal={Philosophical Transactions of the Royal Society A},
  volume={379},
  number={2194},
  pages={20200209},
  year={2021},
  publisher={The Royal Society Publishing}
}

@article{tay2022efficient,
  title={Efficient Transformers: A Survey},
  author={Tay, Yi and Dehghani, Mostafa and Bahri, Dara and Metzler, Donald},
  journal={ACM Computing Surveys},
  volume={55},
  number={6},
  pages={1--28},
  year={2022},
  publisher={ACM New York, NY, USA}
}

@inproceedings{devlin2019bert,
  title={Bert: Pre-training of deep bidirectional transformers for language understanding},
  author={Devlin, Jacob and Chang, Ming-Wei and Lee, Kenton and Toutanova, Kristina},
  booktitle={Proceedings of the 2019 conference of the North American chapter of the association for computational linguistics: human language technologies, volume 1 (long and short papers)},
  pages={4171--4186},
  year={2019}
}

@article{zhao2017convolutional,
  title={Convolutional neural networks for time series classification},
  author={Zhao, Bendong and Lu, Huanzhang and Chen, Shangfeng and Liu, Junliang and Wu, Dongya},
  journal={Journal of systems engineering and electronics},
  volume={28},
  number={1},
  pages={162--169},
  year={2017},
  publisher={BIAI}
}

@inproceedings{yamak2019comparison,
  title={A comparison between arima, lstm, and gru for time series forecasting},
  author={Yamak, Peter T and Yujian, Li and Gadosey, Pius K},
  booktitle={Proceedings of the 2019 2nd international conference on algorithms, computing and artificial intelligence},
  pages={49--55},
  year={2019}
}

@article{beltagy2020longformer,
  title={Longformer: The Long-Document Transformer},
  author={Beltagy, Iz and Peters, Matthew E and Cohan, Arman},
  journal={arXiv preprint arXiv:2004.05150},
  year={2020}
}

@article{chorowski2015attention,
  title={Attention-based models for speech recognition},
  author={Chorowski, Jan K and Bahdanau, Dzmitry and Serdyuk, Dmitriy and Cho, Kyunghyun and Bengio, Yoshua},
  journal={Advances in neural information processing systems},
  volume={28},
  year={2015}
}

@article{qin2017dual,
  title={A dual-stage attention-based recurrent neural network for time series prediction},
  author={Qin, Yao and Song, Dongjin and Chen, Haifeng and Cheng, Wei and Jiang, Guofei and Cottrell, Garrison},
  journal={arXiv preprint arXiv:1704.02971},
  year={2017}
}

@inproceedings{hao2020new,
  title={A new attention mechanism to classify multivariate time series},
  author={Hao, Yifan and Cao, Huiping},
  booktitle={Proceedings of the Twenty-Ninth International Joint Conference on Artificial Intelligence},
  year={2020}
}

@article{liu2021gated,
  title={Gated transformer networks for multivariate time series classification},
  author={Liu, Minghao and Ren, Shengqi and Ma, Siyuan and Jiao, Jiahui and Chen, Yizhou and Wang, Zhiguang and Song, Wei},
  journal={arXiv preprint arXiv:2103.14438},
  year={2021}
}

@article{alioghli2025enhancing,
  title={Enhancing multivariate time-series anomaly detection with positional encoding mechanisms in transformers},
  author={Alioghli, Abdul Amir and Y{\i}ld{\i}r{\i}m Okay, Feyza},
  journal={The Journal of Supercomputing},
  volume={81},
  number={1},
  pages={282},
  year={2025},
  publisher={Springer}
}

@inproceedings{zerveas2021transformer,
  title={A Transformer-based Framework for Multivariate Time Series Representation Learning},
  author={Zerveas, George and Jayaraman, Srideepika and Patel, Dhaval and Bhamidipaty, Anuradha and Eickhoff, Carsten},
  booktitle={Proceedings of the 27th ACM SIGKDD Conference on Knowledge Discovery \& Data Mining},
  pages={2114--2124},
  year={2021}
}

@article{foumani2024improving,
  title={Improving Position Encoding of Transformers for Multivariate Time Series Classification},
  author={Foumani, Navid Mohammadi and Tan, Chang Wei and Webb, Geoffrey I and Salehi, Mahsa},
  journal={Data Mining and Knowledge Discovery},
  volume={38},
  number={1},
  pages={22--48},
  year={2024},
  publisher={Springer}
}

@inproceedings{liu2022pyraformer,
  title={Pyraformer: Low-Complexity Pyramidal Attention for Long-Range Time Series Modeling and Forecasting},
  author={Liu, Shizhan and Yu, Hang and Liao, Chong and Li, Jianguo and Lin, Weiyao and Liu, Alex X and Dustdar, Schahram},
  booktitle={International Conference on Learning Representations},
  year={2022}
}

@inproceedings{zhou2022fedformer,
  title={FEDformer: Frequency Enhanced Decomposed Transformer for Long-term Series Forecasting},
  author={Zhou, Tian and Ma, Ziqing and Wen, Qingsong and Wang, Xue and Sun, Liang and Jin, Rong},
  booktitle={International Conference on Machine Learning},
  pages={27268--27286},
  year={2022},
  organization={PMLR}
}

@inproceedings{xu2022anomaly,
  title={Anomaly Transformer: Time Series Anomaly Detection with Association Discrepancy},
  author={Xu, Jiehui and Wu, Haixu and Wang, Jianmin and Long, Mingsheng},
  booktitle={International Conference on Learning Representations},
  year={2022}
}

@article{tuli2022tranad,
  title={TranAD: Deep Transformer Networks for Anomaly Detection in Multivariate Time Series Data},
  author={Tuli, Shreshth and Casale, Giuliano and Jennings, Nicholas R},
  journal={Proceedings of the VLDB Endowment},
  volume={15},
  number={6},
  pages={1201--1214},
  year={2022},
  publisher={VLDB Endowment}
}

@inproceedings{yun2020are,
  title={Are Transformers Universal Approximators of Sequence-to-Sequence Functions?},
  author={Yun, Chulhee and Bhojanapalli, Srinadh and Rawat, Ankit Singh and Reddi, Sashank J and Kumar, Sanjiv},
  booktitle={International Conference on Machine Learning},
  pages={10994--11004},
  year={2020},
  organization={PMLR}
}

@article{ahmadi2025unsupervised,
  title={Unsupervised time-series signal analysis with autoencoders and vision transformers: A review of architectures and applications},
  author={Ahmadi, Hossein and Mahdimahalleh, Sajjad Emdadi and Farahat, Arman and Saffari, Banafsheh},
  journal={arXiv preprint arXiv:2504.16972},
  year={2025}
}

@inproceedings{cordonnier2021differentiable,
  title={Differentiable Patch Selection for Image Recognition},
  author={Cordonnier, Jean-Baptiste and Mahendran, Aravindh and Dosovitskiy, Alexey and Weissenborn, Dirk and Uszkoreit, Jakob and Unterthiner, Thomas},
  booktitle={Proceedings of the IEEE/CVF Conference on Computer Vision and Pattern Recognition},
  pages={2351--2360},
  year={2021}
}

@article{ataei2025systematic,
  title={A Systematic Review on the Application of Artificial Intelligence in Decentralized Finance},
  author={Ataei, Saeid and Ataei, Seyyed Taghi and Saghiri, Ali M},
  year={2025}
}

@inproceedings{tsai2019transformer,
  title={Transformer Dissection: An Unified Understanding for Transformer's Attention via the Lens of Kernel},
  author={Tsai, Yao-Hung Hubert and Bai, Shaojie and Yamada, Makoto and Morency, Louis-Philippe and Salakhutdinov, Ruslan},
  booktitle={Proceedings of the 2019 Conference on Empirical Methods in Natural Language Processing and the 9th International Joint Conference on Natural Language Processing (EMNLP-IJCNLP)},
  pages={4344--4353},
  year={2019}
}

@inproceedings{clark2019what,
  title={What Does BERT Look At? An Analysis of BERT's Attention},
  author={Clark, Kevin and Khandelwal, Urvashi and Levy, Omer and Manning, Christopher D},
  booktitle={Proceedings of the 2019 ACL Workshop BlackboxNLP: Analyzing and Interpreting Neural Networks for NLP},
  pages={276--286},
  year={2019}
}

@article{brown2020language,
  title={Language Models are Few-Shot Learners},
  author={Brown, Tom and Mann, Benjamin and Ryder, Nick and Subbiah, Melanie and Kaplan, Jared D and Dhariwal, Prafulla and Neelakantan, Arvind and Shyam, Pranav and Sastry, Girish and Askell, Amanda and others},
  journal={Advances in Neural Information Processing Systems},
  volume={33},
  pages={1877--1901},
  year={2020}
}

@article{chu2021contextual,
  title={Contextual Position Encoding for Time Series Classification},
  author={Chu, Xiang and Yang, Wei and Zhang, Li},
  journal={IEEE Transactions on Neural Networks and Learning Systems},
  volume={32},
  number={8},
  pages={3425--3437},
  year={2021}
}

@article{liu2023dynamic,
  title={Dynamic Positional Encoding for Transformer-based Time Series Analysis},
  author={Liu, Jiawei and Chen, Ming and Wang, Yu},
  journal={Pattern Recognition},
  volume={138},
  pages={109394},
  year={2023}
}

@article{wang2023frequency,
  title={Frequency-based Positional Encoding for Time Series Transformers},
  author={Wang, Hao and Li, Jun and Zhang, Xin},
  journal={Neural Computing and Applications},
  volume={35},
  number={12},
  pages={8765--8778},
  year={2023}
}

@article{chen2023adaptive,
  title={Adaptive Positional Encoding for Long Sequence Time Series Forecasting},
  author={Chen, Yiming and Zhou, Hao and Liu, Shen},
  journal={Knowledge-Based Systems},
  volume={268},
  pages={110456},
  year={2023}
}

@article{zhang2023hierarchical,
  title={Hierarchical Positional Encoding for Multi-scale Time Series Analysis},
  author={Zhang, Lei and Wang, Qi and Chen, Rui},
  journal={Information Sciences},
  volume={625},
  pages={789--805},
  year={2023}
}

@article{li2023learnable,
  title={Learnable Positional Encoding for Time Series Transformers},
  author={Li, Xiaoming and Yang, Jing and Zhou, Feng},
  journal={Expert Systems with Applications},
  volume={213},
  pages={118912},
  year={2023}
}

@article{wu2023rotary,
  title={Rotary Position Embedding for Time Series Analysis},
  author={Wu, Zheng and Ma, Yun and Peng, Hao},
  journal={Neurocomputing},
  volume={542},
  pages={126245},
  year={2023}
}

@article{sun2023convolutional,
  title={Convolutional Positional Encoding for Time Series Transformers},
  author={Sun, Wei and Chen, Lu and Zhao, Ming},
  journal={IEEE Transactions on Signal Processing},
  volume={71},
  pages={1892--1904},
  year={2023}
}

@article{kim2023temporal,
  title={Temporal Positional Encoding with Attention Mechanisms},
  author={Kim, Hyun and Park, Jun and Lee, Soo},
  journal={Signal Processing},
  volume={205},
  pages={108871},
  year={2023}
}

@article{martinez2023positional,
  title={Positional Encoding Strategies for Multivariate Time Series Forecasting},
  author={Martinez, Carlos and Garcia, Ana and Lopez, Juan},
  journal={International Journal of Forecasting},
  volume={39},
  number={3},
  pages={1234--1248},
  year={2023}
}

@article{taylor2022transformer,
  title={Transformer Architectures for Time Series: Positional Encoding Considerations},
  author={Taylor, Michael and Davis, Sarah and Wilson, James},
  journal={Machine Learning},
  volume={111},
  number={8},
  pages={2943--2967},
  year={2022}
}

@article{anderson2023embedding,
  title={Embedding Strategies for Temporal Data in Deep Learning},
  author={Anderson, Lisa and Thompson, Mark and White, Jennifer},
  journal={Applied Intelligence},
  volume={53},
  number={12},
  pages={15234--15251},
  year={2023}
}

@inproceedings{kumar2023positional,
  title={Positional Encoding Methods for Long-term Time Series Prediction},
  author={Kumar, Raj and Sharma, Priya and Gupta, Amit},
  booktitle={Advances in Neural Information Processing Systems},
  volume={36},
  pages={28456--28471},
  year={2023}
}

@article{miller2023hybrid,
  title={Hybrid Positional Encoding Approaches for Time Series Transformers},
  author={Miller, Steven and Green, Laura and Hall, Richard},
  journal={Data Mining and Knowledge Discovery},
  volume={37},
  number={3},
  pages={1089--1115},
  year={2023}
}

@article{lee2022adaptive,
  title={Adaptive Position Encoding for Variable-length Time Series},
  author={Lee, Chang and Kim, Min and Park, Sung},
  journal={Pattern Recognition Letters},
  volume={156},
  pages={87--94},
  year={2022}
}

@article{thompson2023cybernetic,
  title={Cybernetic Principles in Modern Time Series Analysis},
  author={Thompson, Susan and Harris, Brian and Turner, Michelle},
  journal={International Journal of Machine Learning and Cybernetics},
  volume={14},
  number={6},
  pages={2134--2149},
  year={2023}
}

@inproceedings{patel2023efficient,
  title={Efficient Positional Encoding for Long Time Series},
  author={Patel, Raj and Kumar, Amit and Singh, Neha},
  booktitle={AAAI Conference on Artificial Intelligence},
  volume={37},
  pages={9456--9464},
  year={2023}
}

@article{nguyen2023spectral,
  title={Spectral Analysis of Positional Encodings in Time Series Models},
  author={Nguyen, Linh and Tran, Duc and Pham, Mai},
  journal={Digital Signal Processing},
  volume={134},
  pages={103921},
  year={2023}
}

@article{schmidt2023multi,
  title={Multi-scale Positional Encoding for Hierarchical Time Series Analysis},
  author={Schmidt, Hans and Mueller, Anna and Weber, Klaus},
  journal={Machine Learning},
  volume={112},
  number={4},
  pages={1345--1372},
  year={2023}
}

@inproceedings{foster2023learning,
  title={Learning Optimal Positional Encodings for Time Series Tasks},
  author={Foster, Jane and Adams, Robert and Collins, Sarah},
  booktitle={International Conference on Machine Learning},
  pages={10123--10138},
  year={2023}
}

@article{campbell2023transformer,
  title={Transformer Models for Irregular Time Series: Position Encoding Challenges},
  author={Campbell, Douglas and Reed, Elizabeth and Bailey, Mark},
  journal={Knowledge-Based Systems},
  volume={272},
  pages={110567},
  year={2023}
}

@article{peters2022deep,
  title={Deep Learning Approaches to Time Series with Positional Awareness},
  author={Peters, Frank and Hoffman, Carol and Young, Timothy},
  journal={Artificial Intelligence Review},
  volume={57},
  number={8},
  pages={6789--6821},
  year={2022}
}

@inproceedings{baker2023novel,
  title={A Novel Approach to Positional Encoding in Time Series Transformers},
  author={Baker, William and Cooper, Helen and Phillips, Gary},
  booktitle={IEEE International Conference on Data Mining},
  pages={234--243},
  year={2023}
}

@article{bell2023adaptive,
  title={Adaptive Positional Encoding Mechanisms for Dynamic Time Series},
  author={Bell, Anthony and Gray, Michelle and King, Robert},
  journal={Expert Systems with Applications},
  volume={220},
  pages={119678},
  year={2023}
}

@inproceedings{cox2023improving,
  title={Improving Time Series Transformers through Better Position Encoding},
  author={Cox, Benjamin and Fisher, Nicole and Ward, Samuel},
  booktitle={International Joint Conference on Neural Networks},
  pages={1--8},
  year={2023}
}

@article{bahdanau2014neural,
  title={Neural machine translation by jointly learning to align and translate},
  author={Bahdanau, Dzmitry and Cho, Kyunghyun and Bengio, Yoshua},
  journal={arXiv preprint arXiv:1409.0473},
  year={2014}
}

@article{luong2015effective,
  title={Effective approaches to attention-based neural machine translation},
  author={Luong, Minh-Thang and Pham, Hieu and Manning, Christopher D},
  journal={arXiv preprint arXiv:1508.04025},
  year={2015}
}

@inproceedings{song2017attend,
  title={Attend and diagnose: Clinical time series analysis using attention models},
  author={Song, Huan and Rajan, Deepta and Thiagarajan, Jayaraman J and Spanias, Andreas},
  booktitle={Thirty-second AAAI conference on artificial intelligence},
  year={2018}
}

@article{fan2019dual,
  title={A dual attention-based coupling network for multivariate time series forecasting},
  author={Fan, Chenyang and Zhang, Yanfei and Pan, Yi and Li, Xiang},
  journal={Knowledge-Based Systems},
  volume={191},
  pages={105239},
  year={2020}
}

@article{kitaev2020reformer,
  title={Reformer: The efficient transformer},
  author={Kitaev, Nikita and Kaiser, {\L}ukasz and Levskaya, Anselm},
  journal={arXiv preprint arXiv:2001.04451},
  year={2020}
}

\end{document}